\documentclass[letterpaper,journal]{IEEEtran}
\newif\ifthesis

\usepackage{graphicx}
\usepackage[linesnumbered,ruled,vlined]{algorithm2e}
\usepackage{subcaption}
\usepackage[linkcolor=blue]{hyperref}
\usepackage{amsmath} 
\usepackage{amssymb}
\usepackage{multirow}
\usepackage{afterpage}
\usepackage{pdflscape}
\usepackage{soul}
\usepackage{bm}
\usepackage{color}
\hypersetup{
                          pdftitle={Robotic Eye-in-hand Visual Servo Axially Align Nasopharyngeal Swabs with the Nasal Cavity},        pdfauthor={Peter Q. Lee, John S. Zelek, and Katja Mombaur},                             colorlinks=true,           linkcolor=red,              citecolor=green,            filecolor=magenta,          urlcolor=cyan,               }

\begin{document}

    \title{Robotic Eye-in-hand Visual Servo Axially Aligning Nasopharyngeal Swabs with the Nasal Cavity}
    \author{Peter Q. Lee$^{1,*}$, John S. Zelek$^1$ and Katja Mombaur$^{1,2,3}$
\thanks{1. Systems Design Engineering, University of Waterloo, 200 University Avenue West, Waterloo, Canada. 2. Mechanical and Mechatronics Engineering, University of Waterloo, 200 University Avenue
  West, Waterloo, Canada. 3. Optimization and Biomechanics for Human-Centred Robotics (BioRobotics Lab), Institute for Anthropomatics and Robotics, Karlsruhe Institute of Technology, Karlsruhe, Germany *pqjlee@uwaterloo.ca  The research presented in this manuscript was first presented in chapter 5 of PL's PhD thesis.}}
\maketitle
\begin{abstract}
  The nasopharyngeal (NP) swab test is a method for collecting cultures to diagnose for different types of respiratory illnesses, including COVID-19. Delegating this task to robots would be beneficial in terms of reducing infection risks and bolstering the healthcare system, but
  a critical component of the NP swab test is having the swab aligned properly with the nasal cavity so that it does not cause excessive discomfort or injury by traveling down the wrong passage. Existing research towards robotic NP swabbing typically assumes the patient's head is held within a fixture. This simplifies the alignment problem, but is also dissimilar to clinical scenarios where patients are typically free-standing. Consequently, our work creates a vision-guided pipeline to allow an instrumented robot arm to properly position and orient NP swabs with respect to the nostrils of free-standing patients. The first component of the pipeline is a precomputed joint lookup table to allow the arm to meet the patient's arbitrary position in the designated workspace, while avoiding joint limits. Our pipeline leverages semantic face models from computer vision to estimate the Euclidean pose of the face with respect to a monocular RGB-D camera placed on the end-effector. These estimates are passed into an unscented Kalman filter on manifolds state estimator and a pose based visual servo control loop to move the swab to the designated pose in front of the nostril. Our pipeline was validated with human trials, featuring a cohort of 25 participants. The system is effective, reaching the nostril for 84\% of participants, and our statistical analysis did not find significant demographic biases within the cohort.
\end{abstract}

\section{Introduction}
\label{sec:vs_intro}

      Nasopharyngeal (NP) swab culture collection is an important procedure in healthcare that is used to diagnose a variety of respiratory illnesses~\cite{Leber2020-173}. The procedure has a healthcare worker (HCW) align a swab next to the nostril, then insert it through the nasal cavity, in a path that moves posteriorly and is superior to the nasal floor, lateral to the nasal septum, medial to the inferior turbinate, and inferior to the cribiform plate, until it reaches the nasopharynx~\cite{Hiebert2021-278}. The cultures from the nasopharynx can then be subjected to biomolecular testing to diagnose illnesses (e.g., the PCR test). During this process, it is critical that the swab is oriented correctly with respect to the nasal cavity. In particular, if the swab is angled too steeply from the transverse plane (pitch angle), it may not reach the nasopharynx and could cause discomfort or injury in the sensitive upper portion of the nasal cavity. Indeed, the medical literature has reported cases where improper technique has resulted in tissue trauma and heavy bleeding~\cite{Fabbris2021-247}.       Uncontrolled insertion angles also create the potential for injury to the brain or nerve clusters near the base of the skull~\cite{Bleier2020-396}.

      Applying robotics to autonomously conduct swabbing in place of HCWs is beneficial for several reasons. First, it provides a way to reduce exposure between HCWs and their potentially infectious patients, a problem that was of particular concern during the COVID-19 pandemic~\cite{12021-435}. Second, it could reduce the variability with how the procedure is performed, which was an observed issue for HCWs~\cite{Hiebert2021-278}. Third, robotics could provide a way to insulate the healthcare system from demographic shifts that could reduce the available workforce of the healthcare system. Because of the massive need for diagnostic testing during the COVID-19 pandemic, and the consequent burden on the healthcare system, there has since been significant interest in applying robotics to upper-respiratory swabbing. The oropharyngeal (OP) swab test is one type of upper-respiratory sampling technique that moves the swab through the open mouth to collect cultures at the back of the throat. There have been some specialized~\cite{Chen2022-391} and nearly complete autonomous solutions developed for OP swabbing~\cite{Sun2023-420}~\cite{Wang2024-423}. From a robotics perspective, the task has the key advantage that the oropharynx is fully observable (unlike the nasopharynx), which allows a combination of vision and force during the swabbing procedure. However, OP swabs have been shown to have less diagnostic sensitivity compared to NP swabs~\cite{Wang2020-397}. Approaches to NP swabbing consist of custom-designed mechanisms~\cite{Wang2020-196}~\cite{Chen2022-313} and more general collaborative manipulator platforms~\cite{Haddadin2024-390} that generally rely on mechanical compliance or force feedback to perform the task. Many of these approaches assume the patient place their heads in fixtures, which simplifies insertion through the nasal cavity in terms of head alignment.

      Consequently, we examine the problem of aligning the swab next to the nostril through the use of an eye-in-hand visual servo system on a collaborative robotic arm, which we refer to as the \textbf{pre-contact phase}.
      Having the correct alignment is a crucial step towards ensuring a successful contact phase, where the swab is inserted into the nasal cavity. This is a result we have previously seen in both simulation~\cite{Lee2022-309} and physical results~\cite{Lee2024-436}. 
      The topic of visual servo of a nasopharyngeal swab was previously explored by~\cite{Hwang2022-310}, which was effective at positioning the swab towards the nostrils of participants held within head fixtures. While the solution was effective for their scenario, there did not appear to be provisions to explicitly control the orientation of the swab if the participant were unrestrained.
      Our work proposes an expanded scenario where the patient is standing unrestrained, rather than having their head placed in a fixture. 
      This more closely resembles the conditions of HCWs performing the test and would be more comfortable for patients, but it also adds additional challenges.
      As previously mentioned, maintaining the correct orientation of the swab relative to the nasal cavity is crucial for performing the test safely.
      Thus, a key function of our system is to robustly identify and estimate the position of the nostrils and the face's orientation relative to the swab from the perspective of a monocular depth camera. These features guide the visual servo system to actuate the arm, directing the swab toward the nostril.
      The system also needs to be capable of adapting to arbitrary positions of the face within the workspace.
      Finally, the system must be robust to the motion from both the arm and the natural head motions of the standing participant.
      In this manuscript we describe our proposed pipeline to meet these requirements, as well as a number of human trials involving participants from different demographic backgrounds
    to both validate the effectiveness of our system and to examine for potential demographic biases in its visual models.

    For the layout of this manuscript, we first examine the literature for estimating the pose of the face and enabling visual servo in Section \ref{sec:vs:background}, which has been crucial for inspiring the methods in the pre-contact phase pipeline. 
    Our proposed pipeline's methods consist of five main groups of components that are described in Section \ref{sec:vs:methods}.
    To meet the constraints imposed by the unrestrained participant's head, we create a joint lookup table to find joint configurations for the arm that place the end-effector near the participant's face, while ensuring the arm could still move around the face without encountering joint limits. A measurement pipeline is designed using a convolutional neural network (CNN)
    to estimate the head orientation and nostril positions~\cite{Guo2020-122} that are also resolved using depth imaging. Stable
    states are retained by applying an unscented Kalman filter on manifolds (UKF-M)~\cite{Brossard2020-191}. The control law of the system
    follows a pose-based visual servo that brings the swab to a desired pose in front of the
    nostril.
    We validate our approach through human trials with participants from a variety of demographic backgrounds in Section \ref{sec:vs:experiments}. The system
    performs well during trials, and we analyze differences in final pose among the population in Section \ref{sec:vs:disc}. We also note methods that can
    be employed to make the system more robust for future trials. Finally, we summarize and conclude the work done in this manuscript in Section \ref{sec:vs:conclusion}.

\section{Background on applied vision}
\label{sec:vs:background}

In this section we provide background on two overarching topics: computer vision models for facial pose estimation using monocular camera images
and the visual servo framework that will actuate the arm during the pre-contact phase. These topics are drawn upon to realize the goal of aligning the swab with the correct orientation and position with respect to the nostril using the feedback from the RGB-D camera.

\subsection{Facial pose estimation}
Identifying facial landmarks and features is a very active area of research in computer vision. While fueled by
applications like biometrics~\cite{SaezTrigueros2018-133} and sentiment analysis~\cite{Wu2019-30}, the developments in identifying lower level features from images are useful to our scenario. Facial pose estimation can typically be divided into two categories of methods \cite{Wu2019-30}, those that fit a template of landmarks based on regions of the face and those that fit a dense 3D model or mesh based on the appearance of the image.

The landmark template approach typically assumes some template of 2D keypoints that correspond to different regions
of the face (e.g., lips, eyes). Most models will typically apply a machine learning model that trains on instances of labeled images to predict the locations of the template in
new images~\cite{Rathod2014-41}.
However, a key weakness to this approach in the context of our scenario is that the template is only retained within the 2D projection of the image, which makes it difficult to recover the 3D pose of the head. While more recent methods attempt to project landmarks into three dimensions \cite{Zadeh2017-56}, the landmarks are sparse, so extrapolation to find arbitrary points of interest on the face (i.e., the nostril) is still a concern.

The dense 3D face modeling approach is generally more flexible and useful to our scenario because we want to obtain the 3D orientation and
positions of arbitrary points like the nostrils on the face. 
Some methods aim to do a full volumetric regression of the entire face~\cite{Jackson2017-123}.
Other methods may rely on having some 3D morphable model, such as the Basel face model
(BFM)~\cite{Paysan2009-131}, which applied principal component analysis (PCA) to a sample of point clouds from laser scans of faces to allow
them to be described by a lower dimensional subspace vector. Guo \textit{et al.}~\cite{Guo2020-122} created the 3DDFA\_V2 model that uses a lightweight
neural network based on the MobileNet architecture~\cite{Howard2019-207} to directly regress BFM and the rotation matrices to correspond the facial vertices with the input image.  As we will explain in Section \ref{sec:TDDFA_V2}, we ultimately chose to integrate 3DDFA\_V2 into our pipeline.

\subsection{Visual servo control}

Visual servo can be described as the synthesis of computer vision and control theory. Several different forms exist,
but generally the camera and a vision pipeline provides a measured feature that is used by an underlying control law.
The control law formulates an error term between the measurement and the goal, which is driven to zero with actuation commands. Generally, the two main flavors are image-based visual servo (IBVS) and
pose based visual servo (PBVS), which is also sometimes called position based visual servo~\cite{Chaumette2006-326}. In IBVS, the control problem is typically formulated to align some designated features within the 2D image plane, while in PBVS, data from the image is projected into features within 3D Cartesian space. The benefits and drawbacks of both approaches vary greatly depending on the specifics of the problem.

Applications of visual servo to robotic swabbing is more limited, but there
are a few works of relevance.
Chen \textit{et al.}~\cite{Chen2022-391} proposed an oropharyngeal swab system solution that uses a compliant mechanism with a visual module
that uses a deep learning framework to segment the pharyngeal wall. Subsequently, the compliant swab is moved to reach the tonsils through the mouth.
Li \textit{et al.}~\cite{Li2023-343} proposed a visual servo system for NP swabs surrounding a ``hierarchical decision framework'' for placing a grasped swab
next to a nostril.  Despite this, it is unclear what methodology is used to visually estimate the state of the nose or
what advantages the approach has beyond doing inverse kinematics.
Most relevant is the work by Hwang \textit{et al.}~\cite{Hwang2022-310} who developed a visual servo system to position swabs next to a participant's nostrils. The authors
use an eye-to-hand (where the camera is placed separately from the manipulator) visual servo system and use IBVS to position the swab next to the nostril of a subject held in a head
fixture. Ultimately, our work takes a different approach by using an eye-in-hand (where the camera is placed on the end-effector) PBVS approach that we believe will be
beneficial to handling the arbitrary poses of the unrestrained, freestanding patients in our experiments. Also, we actively regulate the angle the swab
approaches the nostril at, something that would have a major impact on swab insertion according to our results from previous work~\cite{Lee2022-309}.

\section{Methods}
\label{sec:vs:methods}
\begin{figure}\centering
  \includegraphics[width=0.8\linewidth]{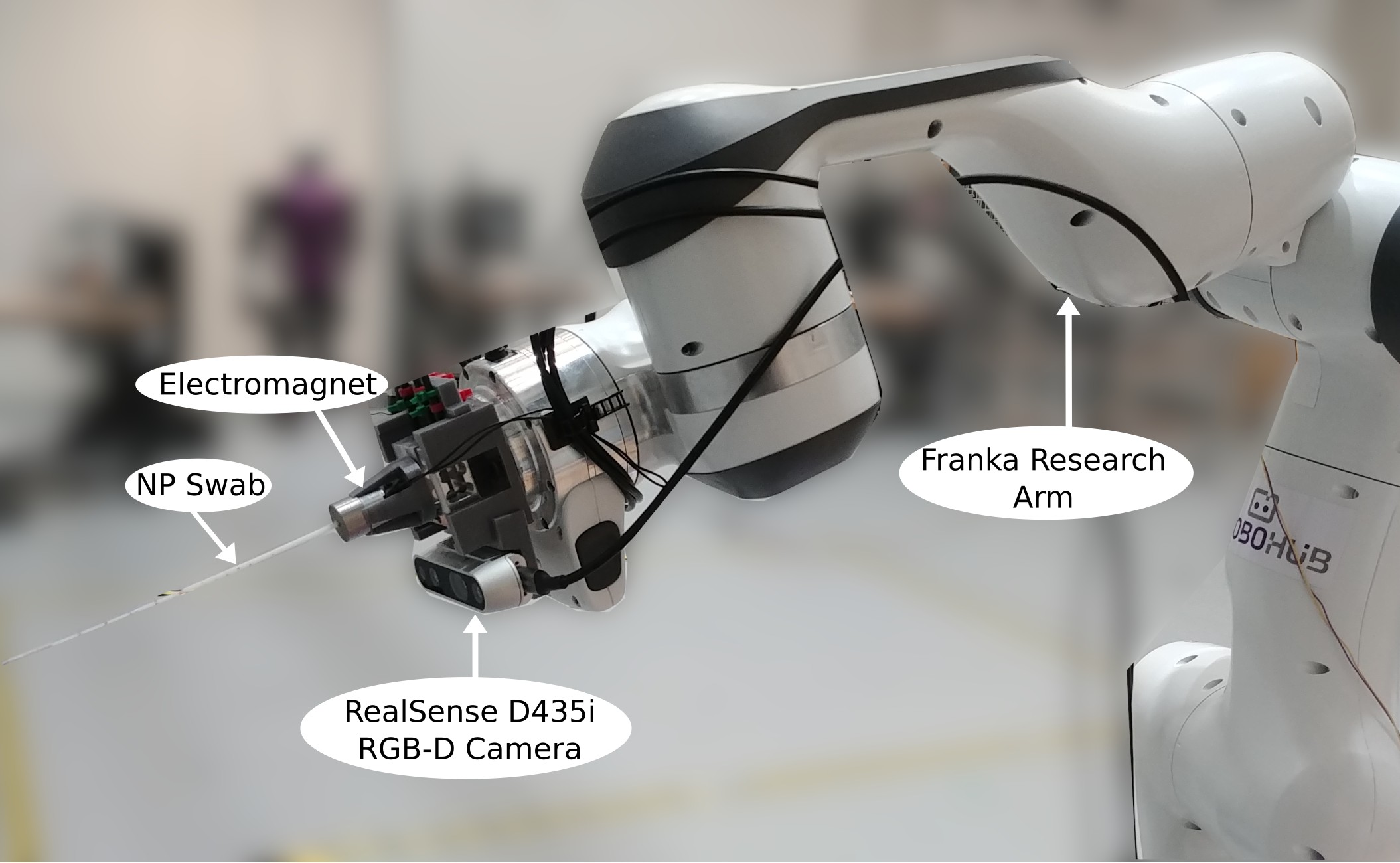}
  \caption{Robotic arm and attached end-effector, equipped with a camera and an electromagnetic mechanism for grasping an NP swab.}
  \label{fig:vs_hardware}
\end{figure}

The main robotic platform used in this work is the Franka Emika research arm, which is a collaborative manipulator with 7 degrees of freedom. We exchanged the standard gripper with a 3D-printed end-effector, which is shown in Fig. \ref{fig:vs_hardware}.
The end-effector was designed under the philosophy that it could generalize to different types of close-contact tasks. Of particular interest for this research is the Intel RealSense D435i RGB-D camera mounted to the end-effector to support visual servo operations and an electromagnet mechanism to grasp tools; in this case, a stainless-steel cylinder containing a NP swab. Note that the camera is oriented so that the swab is within view.

\begin{figure}
      \includegraphics[width=\linewidth]{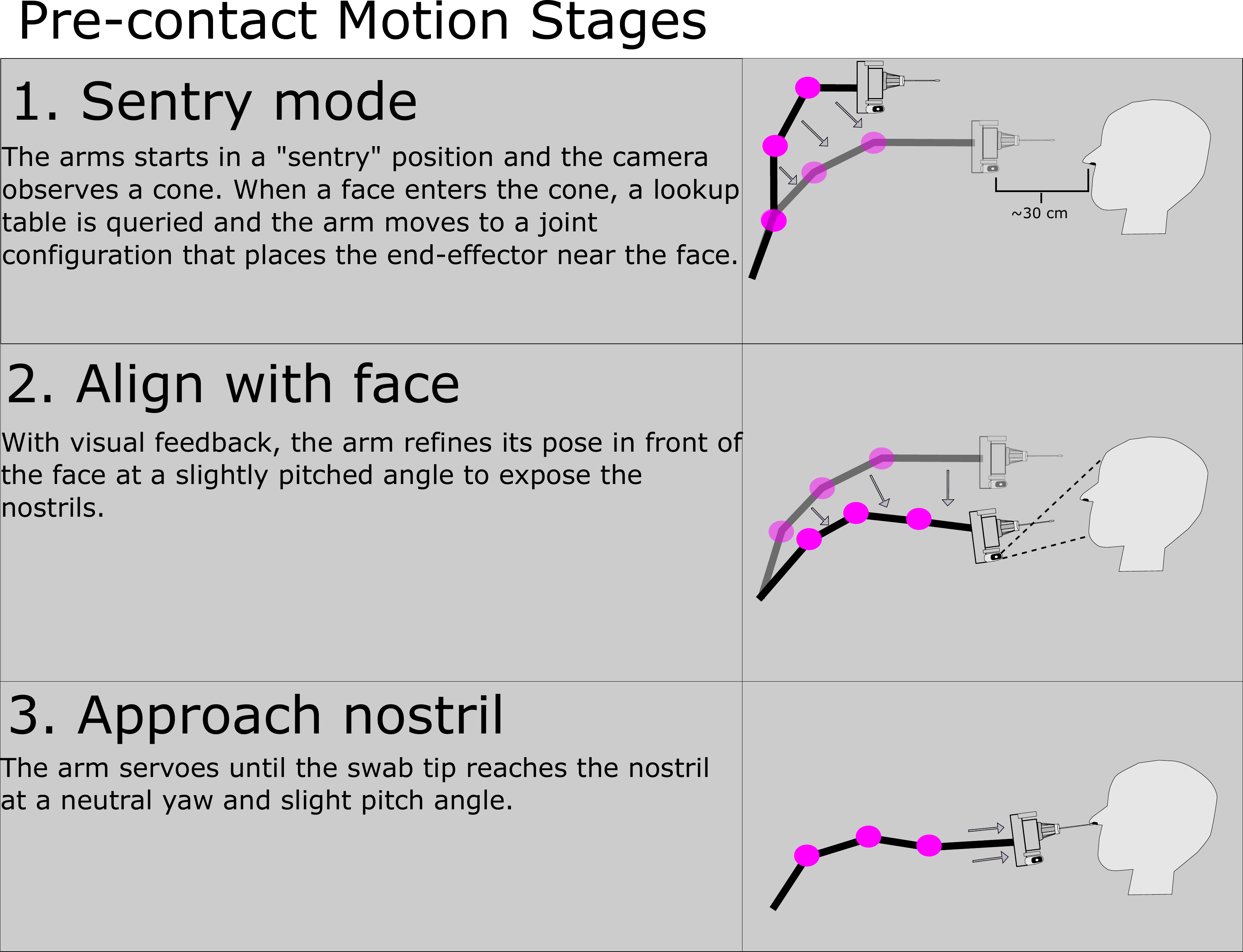}
\caption[Overview of pre-contact sequencing]{Overview of the stages of motion during the pre-contact phase of the NP swab test.}
\label{fig:motion_stages}
\end{figure}

\begin{figure}
  \begin{center}
    \includegraphics[width=\linewidth]{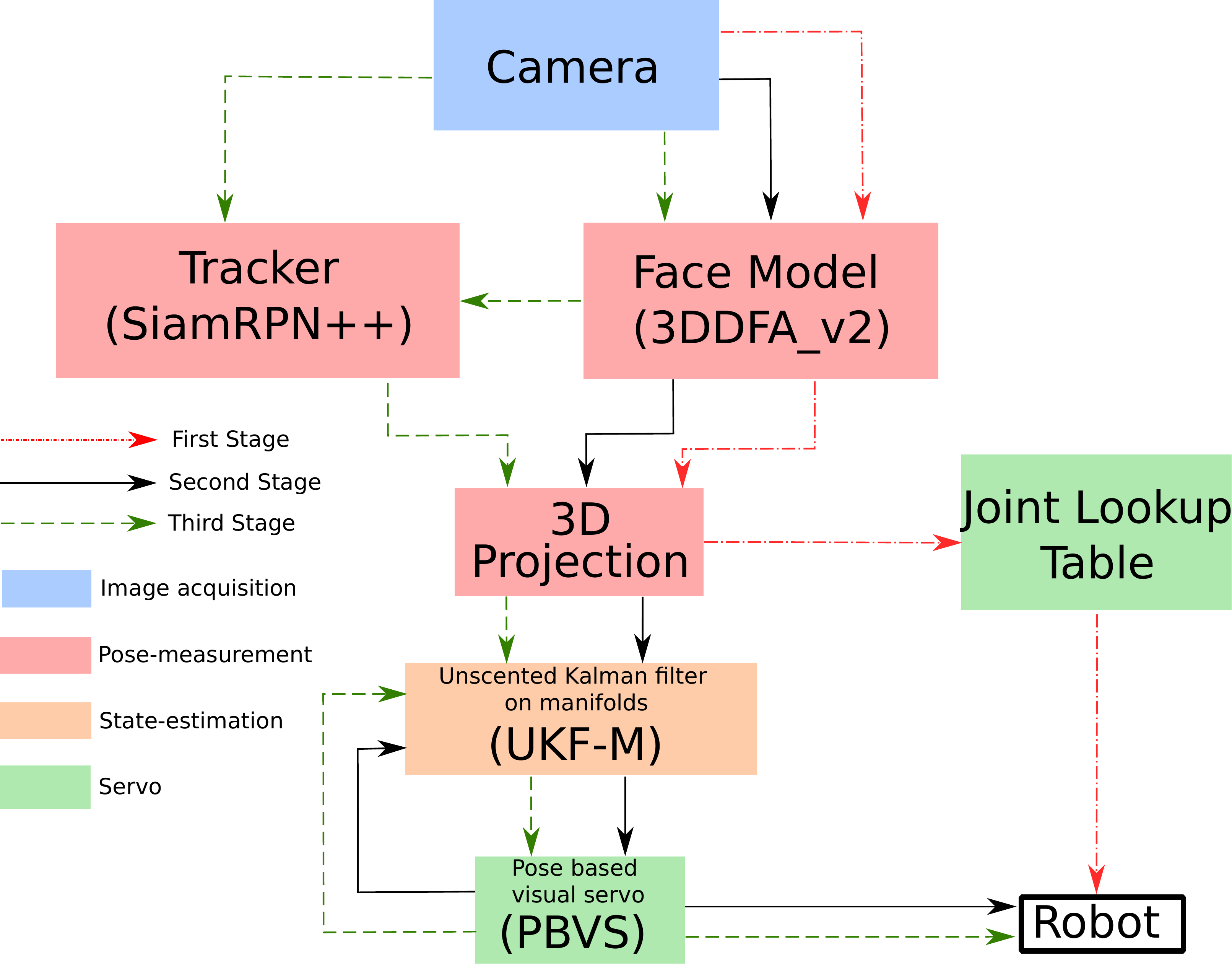}
  \end{center}
  \caption[Flowchart of visual servo system]{Flowchart of the components used throughout the three stages of the pre-contact phase.}
  \label{fig:components}
\end{figure}

We divide the pre-contact phase into three distinct stages of motion, as shown in
Fig. \ref{fig:motion_stages}. During the first stage, the arm begins in a sentry position where
the end-effector camera is oriented to observe the workspace in front of the arm. If a face is detected, a lookup
table is queried and returns a joint configuration that places the end-effector near the face. The second
stage positions the camera 30 cm away (and the swab tip approximately 10 cm away) from
the face at a camera frame pitch of 0.2 radians in order to expose the nostrils. The third stage moves the tip of the
swab to the nostril, at a neutral yaw and 0.2 radian pitch angle.
The system utilizes a face alignment CNN and 3D projection to measure the 3D pose of the face by use of an RGB image and depth image. During the second and third stages, these measurements are fed into a UKF-M state estimator filter to attenuate noise and integrate commanded velocities. The third stage differs in its usage of an unsupervised image segmentation model, and a single object tracker (SOT) CNN to accurately track the nostrils during the final motions. A PBVS control loop computes command velocities for the arm based on the filtered estimates and was implemented with the ViSP framework~\cite{Marchand2005-347}. The interaction of all of these components is ultimately portrayed in Fig.~\ref{fig:components}. The controller code was primarily written in C++, but was extended with Python to run the deep learning models in the visual pipeline and to run the UKF-M.

\subsection{Joint Lookup Table}
\label{sec:sentry}
In this section we describe the resolution of the inverse-kinematics (IK) problem of finding an ideal joint
configuration, $\mathbf{q}_{\text{start}}$, that places the end-effector in proximity to a face located somewhere within the arm's workspace.
Ideally, with joint configuration $\mathbf{q}_{\text{start}}$, the robot would be capable of moving around the face at various directions without
encountering joint limits or self-collisions. 
While there are many point-to-point IK algorithms for redundant manipulators available from the literature (e.g.,
RRT-Connect~\cite{Kuffner2000-291}), running them online without causing a significant delay to the motion sequence remains a concern. In addition, we
would need to extend the methods to cover the area of possible target destinations we expect the arm to move to, rather than just move to a
single destination point, because there will be some uncertainty in the direction the participant is facing.
Therefore, we use a lookup table (LUT) that maps a target input pose to a suitable joint
configuration that is generated with an empirically driven approach.
The lookup table is convenient to use during runtime because it can be queried
extremely fast by finding the table entry that is closest to the detected 3D position of the face via indexing an array. Our algorithm 
for constructing this lookup table is detailed
below.

\begin{figure}
  \includegraphics[width=\linewidth]{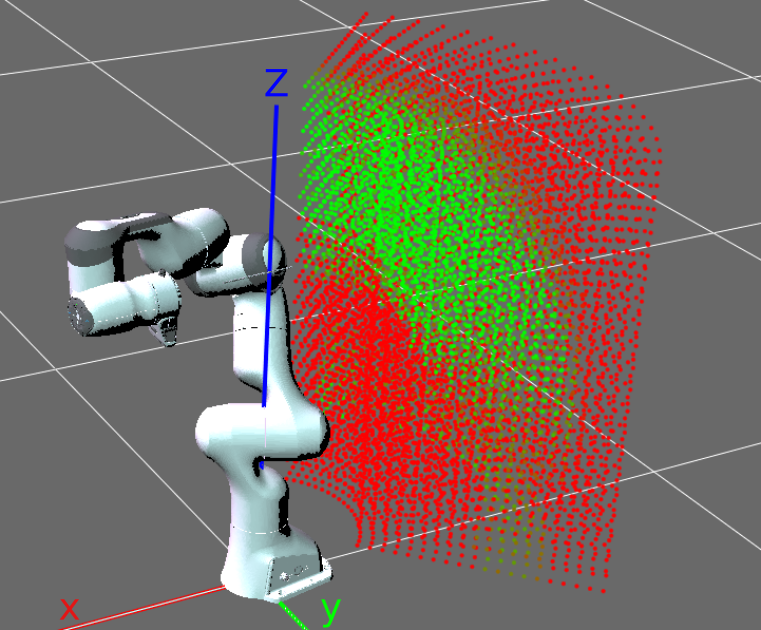}
  \caption[Joint lookup table workspace]{Joint lookup table that spans the arm's workspace cone $\mathcal{C}_{\text{start}}$ (\ref{eq:cone_start}) that represents different possible positions a patient's face can be located at. The color
    represents the achieved quality of the joint configuration stored in the lookup table. The joint positions for green points can reach a high number of points in the destination $\mathcal{C}_{\text{end}}$, while red points reach a low number of points or are infeasible to reach.
  }
  \label{fig:LUT}
  \end{figure}

The sampled points covered by the LUT are contained in a cone, $\mathcal{C}_{\text{start}}\in \text{SE(3)}$, that extends from the field of view of
the end-effector camera at the sentry position. Specifically,
\begin{equation}  
  \mathcal{C}_{\text{start}} = \left\{ \begin{array}{c} (x,y,z,\theta_x,\theta_y,\theta_z), \text{ such that }\\
                                          z_{\text{min}} \le z \le z_{\text{max}},\ r_{\text{min}}\le r \le r_{\text{max}} \\
                                          x = r\cos(\phi),\ y = r\sin(\phi),\ \phi_{\text{min}} \le \phi \le \phi_{\text{max}}\\
                                         \theta_x = 0.2 \text{ rad},\ \theta_y = 0,\ \theta_z = \phi \end{array} \right\}.
                                     \label{eq:cone_start}
\end{equation}
with $\phi_{\text{min}} = -45^\circ$, $\phi_{\text{max}} = 45^\circ$,  $r_{\text{min}} = 0.48$~m, $r_{\text{min}} =
0.68$~m, $z_{\text{min}} = 1$~m, and $z_{\text{max}} = 1.85$~m.
A visual representation of this sampled cone is shown in Fig. \ref{fig:LUT}. 
We can generate different $\mathbf{q}_{\text{start}}$ that place the end-effector in $\mathcal{C}_{\text{start}}$ so that $\bm{\kappa}(\mathbf{q}_{\text{start}}) =
\bm{\epsilon}_{\text{start}}\in\mathcal{C}_{\text{start}}$ by applying the Jacobian pseudo-inverse method
\begin{equation}
  \dot{\mathbf{q}} = \mathbf{J}^{\dagger} (\bar{\bm{\epsilon}} - \bm{\kappa}(\mathbf{q}))
  \label{eq:jpinv_IK}
\end{equation}
where $\bm{\kappa}$ is the forward kinematics function, $\bar{\bm{\epsilon}}$ is the target point for the end-effector, $\mathbf{J}$ is the
robot Jacobian, and ${\mathbf{J}}^\dagger$ is the pseudo-inverse of $\mathbf{J}$. However, given that the Franka arm has joint redundancy, there is no unique solution $\mathbf{q}_{\text{start}}$ for any
given $\bm{\epsilon}_{\text{start}}$~\cite{Lynch2017-139}. As a result, we initialize different candidate joint configurations through sampling from a
random distribution, prior to applying (\ref{eq:jpinv_IK}).

\begin{figure}
  \includegraphics[width=\linewidth]{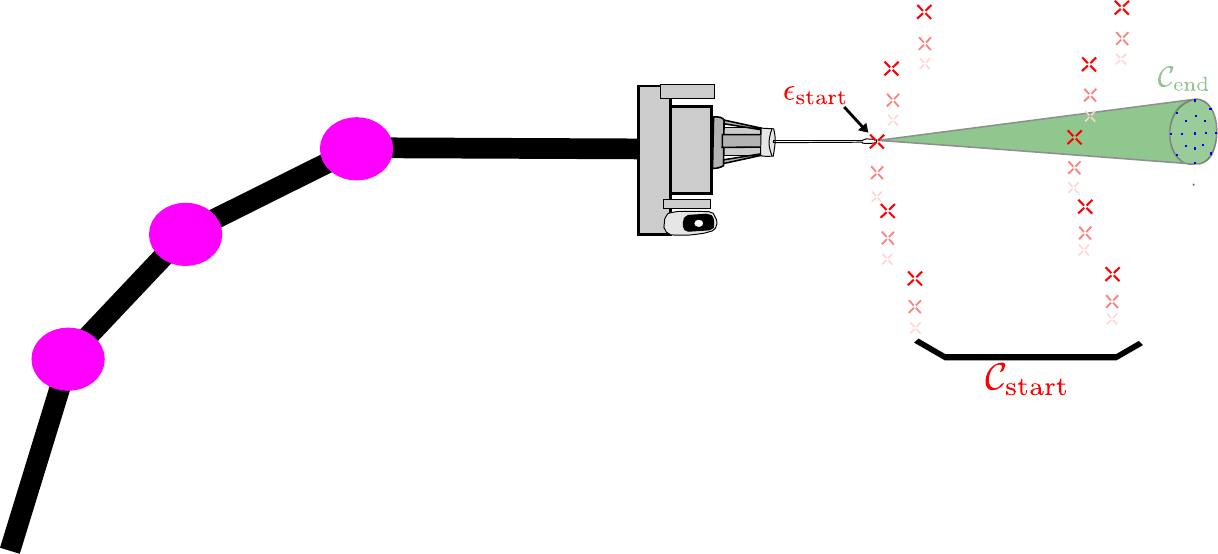}
  \caption[Illustration of cone used to grade workspace]{Illustration of the cone $\mathcal{C}_{\text{end}}$  (\ref{eq:cone_end}) in green extended from a starting point
    $\bm{\epsilon}_{\text{start}}$ (red crosses). During generation of the lookup table, candidate joint configurations are moved to points discretely sampled within this cone (blue dots) to determine how well the arm could move around the face from the starting configuration.}
  \label{fig:extend_cone}
\end{figure}

With a valid set of $\mathbf{q}_{\text{start}}$s, we look to grade the quality of the configurations by attempting to move them to a destination zone.
This destination space takes the form of a second cone
\begin{equation}
  \mathcal{C}_{\text{end}} = \left\{\begin{array}{c} (x,y,z,\theta_x,\theta_y,\theta_z), \text{ such that }\\
                                      r = d\cos(\phi),\  z = d\sin(\phi),\ 0 \le \phi \le \phi_{\text{max}}\\
                                      x = r\cos(\zeta),\  y = r\sin(\zeta), 0 \le \zeta \le
                                      2\pi \\
                                      \theta_{\text{min}} \le \theta_x  \le \theta_{\text{max}},\ \theta_{\text{min}}
                                      \le \theta_y  \le \theta_{\text{max}},\ \theta_z = 0
                                    \end{array}\right\},
                                  \label{eq:cone_end}
                                  \end{equation}
where $d=0.35$ m (distance allocated to reach nostril and insert), $ \phi_{\text{max}} = 15^\circ$, $\theta_{\text{min}} = -10^\circ$, and
$\theta_{\text{max}} = 10^\circ$. For each starting pose $\bm{\epsilon}_{start}$, the points representing the workspace that we
want to sample is $\bm{\epsilon}_{\text{end}} \in \mathcal{T}(\bm{\epsilon}_{start}, \mathcal{C}_{\text{end}})$ where $\mathcal{T}$ is a function that
transforms $\mathcal{C}_{\text{end}}$ to the frame of reference of $\bm{\epsilon}_{\text{start}}$. Fig. \ref{fig:extend_cone} provides a visual
representation of the sampled space extending from $\bm{\epsilon}_{\text{start}}$.
The quality of the candidate $\mathbf{q}_{\text{start}}$ is measured by counting the number ($N$) of $\bm{\epsilon}_{\text{end}} 
\in \mathcal{T}(\bm{\epsilon}_{\text{start}}, \mathcal{C}_{\text{end}})$ that can be reached via (\ref{eq:jpinv_IK}) without encountering any joint limits
or self-collisions. Ultimately, the candidate $\mathbf{q}_{\text{start}}$ with the highest $N$ is stored in the LUT for the
given $\bm{\epsilon}_{\text{start}}$. 
A visualization of the workspace surrounding the robot is shown in Fig. \ref{fig:LUT}, with the maximum $N$ for each entry reflected in
shades from green to red.  Our algorithm utilized the Pinocchio library~\cite{Carpentier2019-404}, which enabled us to compute the Jacobian for performing inverse kinematics and to check for self-collisions. While
generating the lookup table took several hours, querying the lookup table during runtime is extremely efficient, taking a few
milliseconds. Since our workspace is well known ahead of time, the LUT is an effective solution to this problem.

\subsection{Face model \& Tracker (RGB image to facial keypoints and rotation)}  
\label{sec:TDDFA_V2}

Based on the available methods in the literature, we chose to use the 3DDFA\_V2 model by Guo \textit{et al.}~\cite{Guo2020-122} as a major
component of our facial pose measurement pipeline. This model was chosen because it is computationally fast to run and provides a way to directly estimate the rotation of the face as well as arbitrary keypoints of the face.
It relies on an underlying Basel face model~\cite{Paysan2009-131} that is built upon a PCA matrix, $\mathbf{W}$, so the 3D vertices of the face $\mathbf{u}$ can be recovered by taking the
linear combination of its low dimensional basis components $\bm{\beta} \in \mathbb{R}^{50}$ plus the mean values $\overline{\mathbf{u}}$.
\begin{equation}
  \mathbf{u} = \overline{\mathbf{u}} + \mathbf{W}\bm{\beta}.
  \label{eq:bfm}
\end{equation}
The backbone MobileNet~\cite{Howard2019-207} of 3DDFA\_V2 regresses $\mathbf{\beta}$ and a camera projection
matrix $\mathbf{P} \in \mathbb{R}^{3 \times 4}$ based on the shape and orientation of the face under the assumption of a weak projection camera model.
Thus, the vertices of the face mesh following a weak projection camera model are found as
\begin{equation}
  \mathbf{g} = \mathbf{P} \begin{bmatrix}\overline{\mathbf{u}} + \mathbf{W}\mathbf{\beta}\\ 1\end{bmatrix},
  \label{eq:3ddfa_v2}
\end{equation}
where $\mathbf{g}$ is the homogeneous vector of coordinates of the vertices projected into the image plane.
This formulation is convenient, because particular points of interest on the face, e.g. the nostril, may be found by selecting
the subset of rows in $\mathbf{W}$ and the subset of $\overline{\mathbf{u}}$ that correspond to the desired vertices in the face mesh.

While 3DDFA\_V2 is useful for extracting semantic information from a single RGB image, its weak camera projection
assumptions mean some additional steps are needed to obtain real world cardinal measurements. As we will explain in the
following Section \ref{sec:3d_proj}, the depth images from the D435i will be used to find the 3D displacement between the
points of interest and the camera.
The projection matrix, $\mathbf{P}$, can be harnessed to obtain an estimate of the rotation. The left
3$\times$3 submatrix, $\mathbf{R}^*$, of $\mathbf{P}$ contains both scale and rotation of the projected face. 
However, $\mathbf{P}$ is produced by an unconstrained regression, so the vectors of the matrix are not orthogonal like a proper rotation matrix. To fix both of these issues, we
apply singular value decomposition to obtain the closest orthogonal matrix, $\widetilde{\mathbf{R}}$, with $\text{det}(\widetilde{\mathbf{R}})=1$~\cite{Levinson2020-218}:
\begin{equation}
  \begin{aligned}
    &\mathbf{R}^*=\mathbf{U} \bm{\Sigma} \mathbf{V}^{\top},\\
    &\widetilde{\mathbf{R}}=\mathbf{U}\bm{\Sigma}^*\mathbf{V}^{\top},\ \bm{\Sigma}^* =  \text{diag}(1,1,\det(\mathbf{UV}^{\top})).
  \end{aligned}
\end{equation}

Some additional compensation is also needed to make the estimated rotation compatible with the pinhole camera model,
which better corresponds to the image projection of the camera than the weak projection model.
For example, if the face and the camera were oriented in parallel directions, but there is a translation off-center
from the principle axis of the camera, the image of the face would be projected at angle and the weak projection image
would assume the face is rotated off-center.
This is not desirable because we want to decouple the translation and rotational degrees of freedom. 
Therefore, we can find the angles between the face and the principal axis,
\begin{equation}
  \phi_x = \text{atan2}(p_y, p_z), \\
    \phi_y = \text{atan2}(p_x, p_z)\\
\end{equation}
where $p_x, p_y, p_z$ are the 3D translation components between the camera center and the center of the face.
So the final rotation matrix is found by applying the appropriate counter rotation,
\begin{equation}
  \mathbf{R} = \exp([\phi^x,\phi^y,0]_\times)^\top \widetilde{\mathbf{R}},
\end{equation}
where $[\mathbf{z}]_\times$ produces the skew-symmetric matrix containing elements from $\mathbf{z} \in \mathbb{R}^{3}$, and $\exp$ is the matrix exponential. This effectively maps the axis angle rotation vector to a rotation matrix.

The 3DDFA\_V2 model is used during stage 2 and stage 3 to obtain orientation estimates and to find the vertex positions of
the nostril via the facial mesh (\ref{eq:3ddfa_v2}) during stage 2 when the end-effector is further from the face.
However, during stage 3 the swab needs to be closer to the nostrils and a more precise location estimate of the
nostrils is needed. Because we lacked a dedicated nostril detector or data train a detector, we rely on
an unsupervised pipeline to track the nostrils during stage 3. At the end of stage 2, the nostrils are
exposed, so we apply a single object tracking (SOT) network, SiamRPN++~\cite{Li2019-348}, to track them during subsequent frames. SOT models differ
from typical detection models because they can be applied to track arbitrary objects that the
network has not directly trained to detect. The initial bounding box for the nostrils is passed to initialize the SOT and is found by
projecting a selection of vertices around the nostril from (\ref{eq:3ddfa_v2}), then refining the area of interest with the Segment Anything Model (SAM)~\cite{Kirillov2023-393}. During subsequent frames, the SOT model iterates to find the
updated bounding box. The tracked bounding box location is periodically reset if it wanders too far the vertex positions
estimated by 3DDFA\_V2.

\subsection{3D projection}
\label{sec:3d_proj}
The camera is placed on the end-effector of the arm, and therefore our system will act as an eye-in-hand visual
servo system. This greatly simplifies calibration of the system because calibration only needs to be done once, in contrast
to an eye-to-hand setup, because the camera and the end-effector are rigidly attached together. The calibration procedure
derives the spatial transform between the flange of the robot and the camera ${}^e\mathbf{M}_c$, which allows 
observations in the camera frame to be eventually be transformed back to the joint space for actuation. We computed ${}^e\mathbf{M}_c$ by moving
the end-effector around a calibration checkerboard, then applying Tsai's algorithm~\cite{Tsai1987-394} implemented in ViSP~\cite{Marchand2005-347}.
  
  The model in \autoref{sec:TDDFA_V2} is able to identify the rotation of the face and the position
  of keypoints within the image plane. The next step is to use this information to determine the
  relative pose between the nostril and the swab. The 3D position of the nostril is determined using
  the depth channel of the Intel RealSense D435i camera.
  Thus, the 3D position
  \begin{equation}
    \mathbf{p} = \mathbf{K}^{-1}\begin{bmatrix} c\\r\\1 \end{bmatrix} d(c,r)
  \end{equation}
  is determined by using the camera's intrinsic matrix $\mathbf{K}$ for the nostril keypoint at column $c$ and row $r$, with $d(r,c)$ as the depth value in meters for the given row and column in the image.

\begin{figure}
      \includegraphics[width=\linewidth]{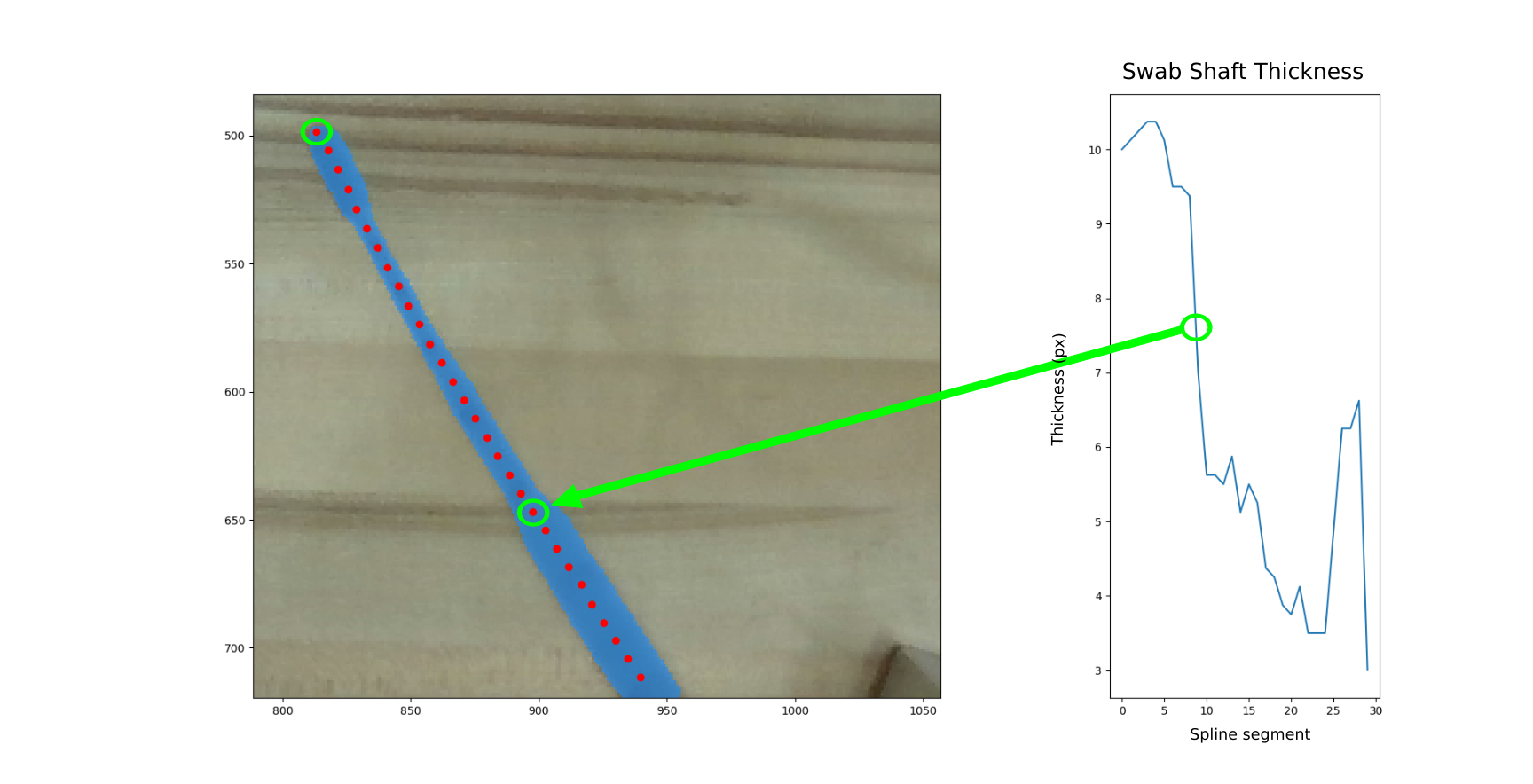}
    \caption[Swab segmentation]{Swab outline is segmented (blue mask) and then fit with a spline through the center (red dots). The shaft point of
      interest ($\mathbf{x}_1$) is found by taking the point belonging to the largest change of thickness within the
      first half of the spline. The tip ($\mathbf{x}_0$) is the final point along the spline.}
    \label{fig:SAM-example}
  \end{figure}

The location and orientation the swab tip with respect to the camera must be known for the visual servo
task. In theory, one could use the measured dimensions of the end-effector schematics to determine the swab's
pose. However, due to placement tolerances and the fact that the swabs have some variable
curvature, it is necessary to estimate the pose of the swab within the camera frame.
   First, before the visual servo process begins, the swab is segmented in the image
  using SAM~\cite{Kirillov2023-393} (an example is shown in Fig. \ref{fig:SAM-example}). The shaft of the swab is parameterized with a spline running
  through the center. The thickness is quantified according to the extent the segmented area covers perpendicular to the spline. This allows us to locate two
  keypoints from the segmentation: $\mathbf{x}_0$, the tip of the swab (146 mm from the end), and $\mathbf{x}_1$, the point where the shaft
  of the swab narrows between 2.5 mm diameter to 1.4 mm diameter (76 mm from the end), which is demarcated by a rapid change in thickness. These two points are used to determine the
  position and direction of the swab with respect to the camera. However, the swab is placed too
  close to the camera to utilize the depth camera. Instead, the cardinal 3D points are estimated using a
  combination of projection from the camera and from the coordinates from the end-effector schematics. 
  Ultimately, we use least squares regression to find the 3D positions of the swab, through ray back-projection 
  \begin{equation}
    \min_{\lambda_0,\lambda_1}{||\lambda_0\mathbf{P}^{\dagger}\mathbf{x}_0 - \tilde{\mathbf{X}}^c_0|| + ||\lambda_1\mathbf{P}^{\dagger}x_1 - \tilde{\mathbf{X}}^c_1||} .
  \end{equation}
  where $\lambda_0, \lambda_1$ are two ray lengths projected from the camera, $\mathbf{P}$ is the camera projection matrix, and
  $\tilde{\mathbf{X}}^c_0, \tilde{\mathbf{X}}^c_1$ are the 3D points on the swab estimated through the schematics.
    With the two points $\mathbf{X}^c_0 = \lambda_0\mathbf{P}^{\dagger}\mathbf{x}_0 $ and $\mathbf{X}^c_1 = \lambda_1\mathbf{P}^{\dagger}\mathbf{x}_1$
    calculated, the 3D position of the swab tip and its direction vector, $\frac{\mathbf{X}^c_0 - \mathbf{X}^c_1}{||\mathbf{X}^c_0 - \mathbf{X}^c_1||}$, define the current pose of the swab.
    This process only needs to be done once per trial because the swab and camera undergo the same
    motion.
    
  The final step of the pose estimation system is to determine the pose difference between the swab and the desired pose
  in front of the nostril. 
  The relative position can be found by simply subtracting the position of the
    nostril from the swab tip $\mathbf{p} - \mathbf{X}^c_0$. During stage 2, we have the desired rotation of the camera $\mathbf{R}_d$ set at 0.2
    radians, so the relative rotation is then $\mathbf{R}_d\mathbf{R}^{\top}$. For stage 3, the swab does not need to
    be oriented with all three rotational degrees of freedom; only the swab's direction vector and
    the desired normal vector of the nostril need to align. To find the ``minimal'' rotation to align these two
    vectors, suppose the vector of the swab direction is $\mathbf{v}$ and the target
    normal of the nostril is $\mathbf{n}$, pitched at 0.2 radians. Then the relative rotation $\mathbf{R}$ can be found as~\cite{httpsmathstackexchangecomusers91768jurvandenberg2013-405}:
    \begin{equation}
      \begin{aligned}
        \mathbf{w} &= \mathbf{v} \times \mathbf{n},\ c = \mathbf{v} \cdot \mathbf{n}\\
        \mathbf{R} &= \mathbf{I}_3 + [\mathbf{w}]_\times  + [\mathbf{w}]_\times [\mathbf{w}]_\times \left(\frac{1}{1+c}\right)
      \end{aligned}
      \label{eq:min_z_rotation}
    \end{equation}
    with $\times$ as the cross product operator and $\mathbf{I}_n$ as the $n\times n$ identity matrix.

\subsection{Unscented Kalman filter on manifolds (UKF-M)}
\label{sec:ukfm}

Our initial evaluations of the measurement pipeline found significant noise in
the pose measurement system that would impede the performance of the visual
servo, if left uncompensated. Another concern is that if measurements are lost from depth acquisition or failure from the face model, then the
servo would intermittently halt. Therefore, 
having a state estimator in the system is useful to compensate for these sources of noise and enable the control loop to continuously run. 
Since the measured state is in the 3D Euclidean space group (SE(3)) a non-standard Kalman filter will be required
because of non-linear dynamics. In addition, classical Kalman filters~\cite{Kalman1960-406} assume the state as a vector, which is problematic because
using vector representations of rotation can instigate wrap around and gimbal lock issues. Therefore, we
used an unscented Kalman filter on manifolds~\cite{Brossard2020-191} to seamlessly estimate the state within this domain. First, the process equations of motion are defined as
\begin{equation}
  \begin{aligned}
    \mathbf{p}(t+\Delta t) &= \mathbf{p}(t) + \Delta t (-\mathbf{v}_p - \mathbf{v}_\theta \times \mathbf{p}(t))\\
   \mathbf{R}(t+\Delta t) &= \mathbf{R}(t) \exp(\Delta t [-\mathbf{v}_\theta]_\times)
    \end{aligned}
  \label{eq:eq_state_trans}
\end{equation}
where $\mathbf{p} \in \mathbb{R}^3$ is the position, $\mathbf{R} \in\ \text{SO(3)}$ is the rotation matrix, $\mathbf{v}_p \in \mathbb{R}^3$ and $\mathbf{v}_\theta \in \mathbb{R}^3$ are the linear and angular velocity of the camera frame that is commanded by the control
loop, and $\Delta t$ is the time step.

The UKF-M contains two innovations over the standard Kalman filter: a manifold projection to handle the non-vector state, and the unscented
transform to handle the non-linear dynamics. 
Manifold projection requires us to define two functions $\phi$ and $\phi^{-1}$ that correspond to our specific state.
The retraction function $\phi(X,\xi) \in$ SE(3) makes an incremental change to the state $X\in $ SE(3) by a
perturbation vector $\xi \in \mathbb{R}^6$. For the translation component, this amounts to vector addition (e.g., $\phi(p,\xi_p)
= \mathbf{p}+\xi_p$) while the rotational component treats $\xi_R \in \mathbb{R}^3$ as an axis-angle rotation vector so $\phi(\mathbf{R},\xi_R) = \mathbf{R} \exp([\xi_R]_\times)$. 
In addition, the inverse-retraction function $\phi^{-1}(X, \hat{X}) \in \mathbb{R}^6$ maps the change between the two SE(3)
states to a 6-D vector. The translation component corresponds to vector subtraction (e.g., $\phi^{-1}(\mathbf{p},\hat{\mathbf{p}}) = \mathbf{p}-
\hat{\mathbf{p}}$) while the rotation component converts the relative rotation between the two matrices into an axis-angle vector
($\phi(\mathbf{R},\hat{\mathbf{R}}) = \log(\mathbf{R} \hat{\mathbf{R}}^{\top})$). 

Like a classical Kalman filter, the UKF-M needs to have an estimate of the statistics of the state (i.e., the mean and covariance) in order to
balance the information between new measurements and process dynamics when updating the state. The difference is that a Kalman filter
assumes a linear process, where the statistics would accurately propagate with linear transformations~\cite{Kalman1960-406}. These assumptions do not
transfer well to a non-linear process~\cite{Julier2004-407}, which is what the unscented transform attempts to rectify.
The unscented transform operates by obtaining deterministic samples perturbed around the state and feeding them through the non-linear propagation function 
to estimate a more accurate the set of statistical parameters~\cite{Julier2004-407}. The UKF-M also adopts this unscented transform to
estimate the state statistics, just by making use of the $\phi$ and $\phi^{-1}$ functions project the perturbed samples into the manifold
and vice versa through the non-linear functions.

The remaining parameters that need to be defined to apply the
UKF-M are the measurement covariance matrix, process covariance matrix, and the
scale parameter for the unscented transform.
Based on the noise characteristics of the pose estimation system (Sections \ref{sec:TDDFA_V2} and \ref{sec:3d_proj}), we hand tuned the measurement covariance components corresponding to translation to $0.005\mathbf{I}_3$ 
and to rotation to $0.05\mathbf{I}_3$. 
The process covariance matrix was set at
$0.01\mathbf{I}_6$. 
The scale parameters were set at 0.01 for state propagation, 0.1 for noise propagation, and 0.01 for
state updates.

\subsection{Pose Based Visual Servo (PBVS)}
\label{sec:PBVS}
Motion of the arm is implemented using a pose based visual servo (PBVS) control law. This component takes the
observed state pose state from the UKF-M and
computes the joint velocities that will move the end-effector to the desired position $\mathbf{t}_d$ and
rotation $\mathbf{R}_d$. The control law~\cite{Chaumette2006-326} in the camera frame takes the form of 
\begin{equation}
  \mathbf{v}_c =  \lambda \mathbf{L}^{\dagger} \mathbf{s}
\end{equation}
where $\mathbf{s}$ is the pose error between the desired pose and the current pose from UKF-M.
The interaction matrix, $\mathbf{L}$, 
is used to map the velocity of the camera origin to the
velocity of the target in the camera's frame. In this vein, the pseudo inverse is used to map
pose error into velocities of the camera. The interaction matrix
is a block matrix $\mathbf{L} = \begin{bmatrix} \mathbf{L}_t & \mathbf{0} \\  \mathbf{0} & \mathbf{L}_{\theta u} \end{bmatrix}$, where the
non-zero blocks correspond to the translational and rotational components of the pose vector. Given that $\mathbf{R}$ is the rotational component of
$\mathbf{s}$ and $ \theta \mathbf{u} = \log(\mathbf{R}),\ ||\mathbf{u}||=1$, these
components are defined as $\mathbf{L}_t=\mathbf{R}$ and $\mathbf{L}_{\theta u} = \mathbf{I}_3 - \frac{\theta}{2} [\mathbf{u}]_\times + (1-
\frac{sinc(\theta)}{sinc^2(\frac{\theta}{2})})[\mathbf{u}]_\times[\mathbf{u}]_\times$~\cite{Chaumette2006-326}.
Finally, the end-effector robot Jacobian in the camera's frame ${}^c\mathbf{J}_{ee}$ is used to
map the desired camera velocity into joint velocities
\begin{equation}
  \mathbf{v}_q = {}^c\mathbf{J}_{ee}^{\dagger} \mathbf{v}_c = \lambda  {}^c\mathbf{J}_{ee}^{\dagger}\mathbf{L}^{\dagger}\mathbf{s}.
\end{equation}
The gain $\lambda$ was kept at a value of 0.5 s$^{-1}$. This control law was implemented using ViSP~\cite{Marchand2005-347} through the C++
Franka Control Interface.

\section{Experiments}\label{sec:vs:experiments}
\begin{figure*}
  \includegraphics[width=\linewidth]{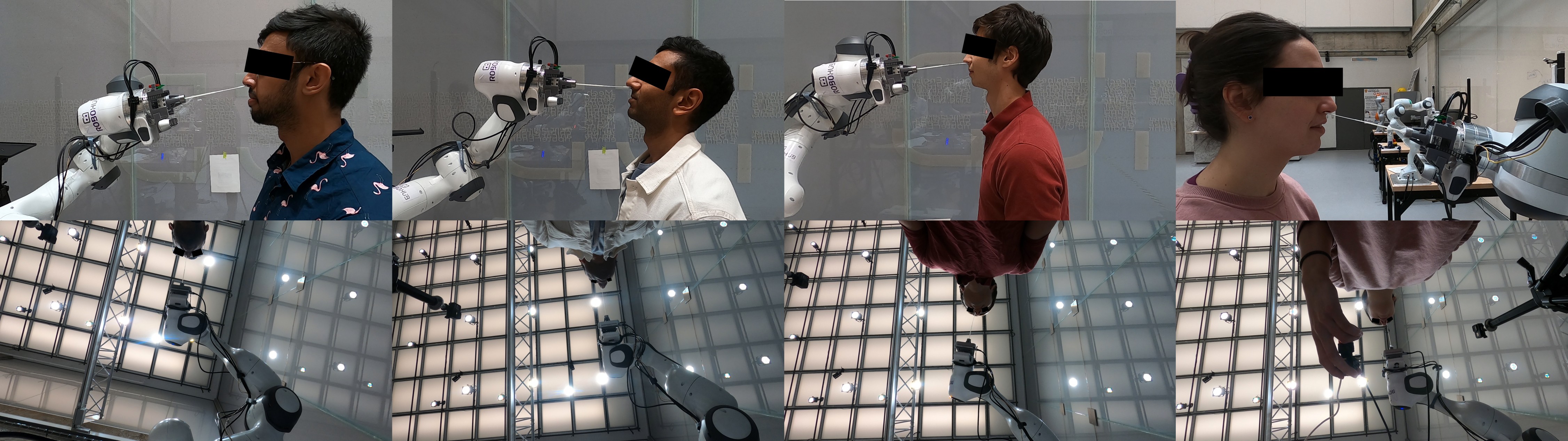}
  \caption[Trial outcome images]{Images showing the trials of several different participants. The images were taken once the swab tip reached the nostril.}
  \label{fig:vs:trial_outcomes}
\end{figure*}

To evaluate the efficacy the proposed pre-contact phase pipeline and its components, we conducted a series of human trials, which were approved by the
University of Waterloo institutional research ethics board (REB \# 43485). The experiments took place from July 2023 to February 2024 at the University of Waterloo RoboHub with a total of 25
participants, who were primarily recruited from mailing lists in the Faculty of Engineering. The
participants were
instructed to freely stand within the arm's cone workspace so the system could move the
swab towards one of their nostrils. This process was repeated for the other nostril. The outcome of the trial was
recorded by two perpendicular GoPro Hero 8 cameras taken from the side and bottom directions. Fig. \ref{fig:vs:trial_outcomes} shows examples of some of the trials using camera arrangement. One question of particular interest is whether the performance of the face pose estimation system is influenced
by population demographics because 3DDFA\_V2 was trained on empirical data that could have under-represented some demographic groups. 
Therefore, participants also filled out a short survey where they could
indicate sex and racial information, which is summarized in Table \ref{tab:demographics}.

\begin{table}
  \caption[Study demographics]{Demographics of study participants that were recorded to evaluate for potential model biases. The mixed race participants are assigned fractionally to each selected race.}
  \label{tab:demographics}
  \begin{center}
    \begin{tabular}{|c|c|c|c|c|c|}\hline
  \multicolumn{6}{|c|}{\textbf{Sex}}\\\hline
  \multicolumn{3}{|c|}{\textbf{Male}} & \multicolumn{3}{|c|}{\textbf{Female}}\\
  \multicolumn{3}{|c|}{17} & \multicolumn{3}{|c|}{8} \\\hline\hline
  \multicolumn{6}{|c|}{\textbf{Race}}\\\hline
  \multirow{2}{*}{\textbf{White}} & \textbf{South} & \textbf{East} & \textbf{Latin} & \multirow{2}{*}{\textbf{Iranian}} &
                                                                                                  \multirow{2}{*}{\textbf{Arab}}\\
  &\textbf{Asian}& \textbf{Asian} & \textbf{American}& &\\
  10.5 & 6.5 & 3.5 & 2 & 2 & 0.5\\ \hline
\end{tabular}

    \end{center}
  \end{table}

\subsection{Positioning and alignment outcomes}

In terms of general results, 
the system was quite successful at reaching the nostril,
although the system did exhibit some discrepancy among the faces encountered.
The system was able to reach the nostril during 84\%\ of the trials. The remaining 16\% of cases
placed the tip of the swab just below the nostril (above the lip). This state generally occurred
because of the failure of 3DDFA\_V2 to locate a good initial bounding box of the nostrils
correctly during stage 3 (see Fig. \ref{fig:missed_nostril}). As a result, the SOT component would be tracking the wrong location
throughout stage 3.  
There was one case where a man with a darker beard caused the SOT tracker to diverge and move the
bounding box towards the beard periodically. In Section \ref{sec:vs_improvements}, we discuss some ideas to rectify these issues.

\begin{figure}
  \begin{subfigure}{0.5\linewidth}
    \begin{center}
      \includegraphics[width=\linewidth]{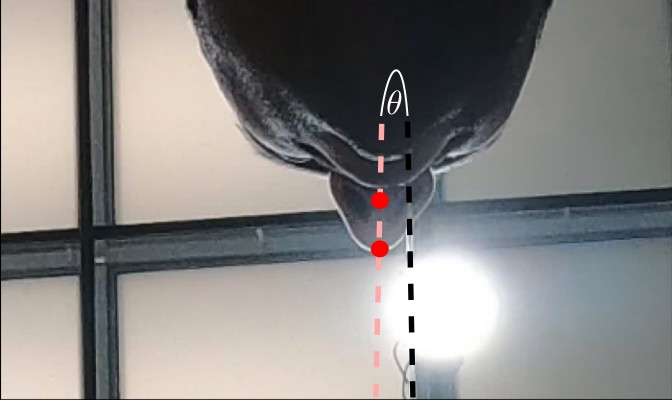}
    \end{center}
    \caption{}
  \end{subfigure}\begin{subfigure}{0.5\linewidth}
    \begin{center}
      \includegraphics[width=\linewidth]{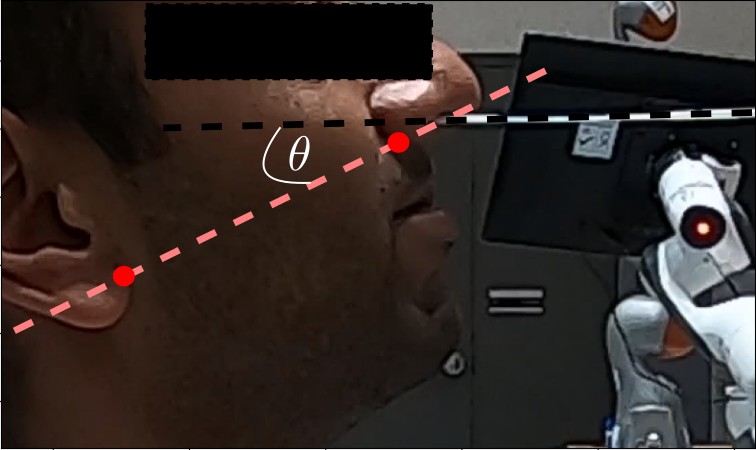}
    \end{center}
      \caption{}
  \end{subfigure}
  \caption[Face labeling scheme]{Image showing the labeling scheme for determining the relative angles between the swab
    and the face. (a) The bottom view landmarks align with the columella of the nose; (b) the
    side view landmarks are the corner of the ear and the ala of the nose (see \cite{279}). The reported angles come from
    the angular difference of the line segments between the landmarks (pink dashed) and the swab (black dashed). 
  }
  \label{fig:face_label}
\end{figure}

While placing the tip of the swab at the nostril is obviously important, the orientation of the swab
is also crucial to regulate to prior to the contact phase. 
To this end, another objective of our experiments is examining the final orientation of the swab
among the differ trials. We estimated the final pitch and yaw orientation within the trials by annotating the
footage of the perpendicular GoPro cameras on the terminal frame. The pitch orientation was found by
comparing the angle difference between the line segment of the swab (from tip to shaft), and the
line segment connecting the point of the earlobe to the Ala (lower edge of the nose) within the side view, as shown in
Fig. \ref{fig:face_label}. The yaw angle took the angular offset between the swab line segment and the line segment from the base of the
columella to the tip of the nose from the bottom view. The distribution of these two quantities is shown in
Fig. \ref{fig:obs_swab_ang_distr}. Yaw had a mean angle of 
-4.09$^\circ$ and standard deviation of 7.82$^\circ$, while pitch had a mean of 16.31$^\circ$, standard deviation of 4.48$^\circ$.

The characteristics of swab to camera pose calibration from Section \ref{sec:3d_proj} is worth examining as well. From the camera's point of view, it was clear that the position of the swab did indeed vary slightly between trials because each swab had slightly different curvature due to the manufacturing process and there were some placement tolerances based on how it was inserted into the steel cylinder. While physically these factors did not appear significant, it is prudent to verify that the process is making reasonable estimates.
Compared to the schematics, there was an average positional offset of 5.2 mm from the center of the two keypoints $\mathbf{X}^c_0$ and $\mathbf{X}^c_1$, and an average angular offset of 3.5$^\circ$. However, the estimated offset in each trial were fairly consistent, as the standard deviation for position was only 1.4 mm and $1.4^\circ$ for position and angle respectively. As a result, it is unlikely that the swab to camera pose estimation is to blame for the variance observed in the previous paragraph, and therefore must be attributed to other sources.

\subsection{Noise attenuation}
\label{sec:noise_att}
\begin{figure}
  \includegraphics[width=\linewidth]{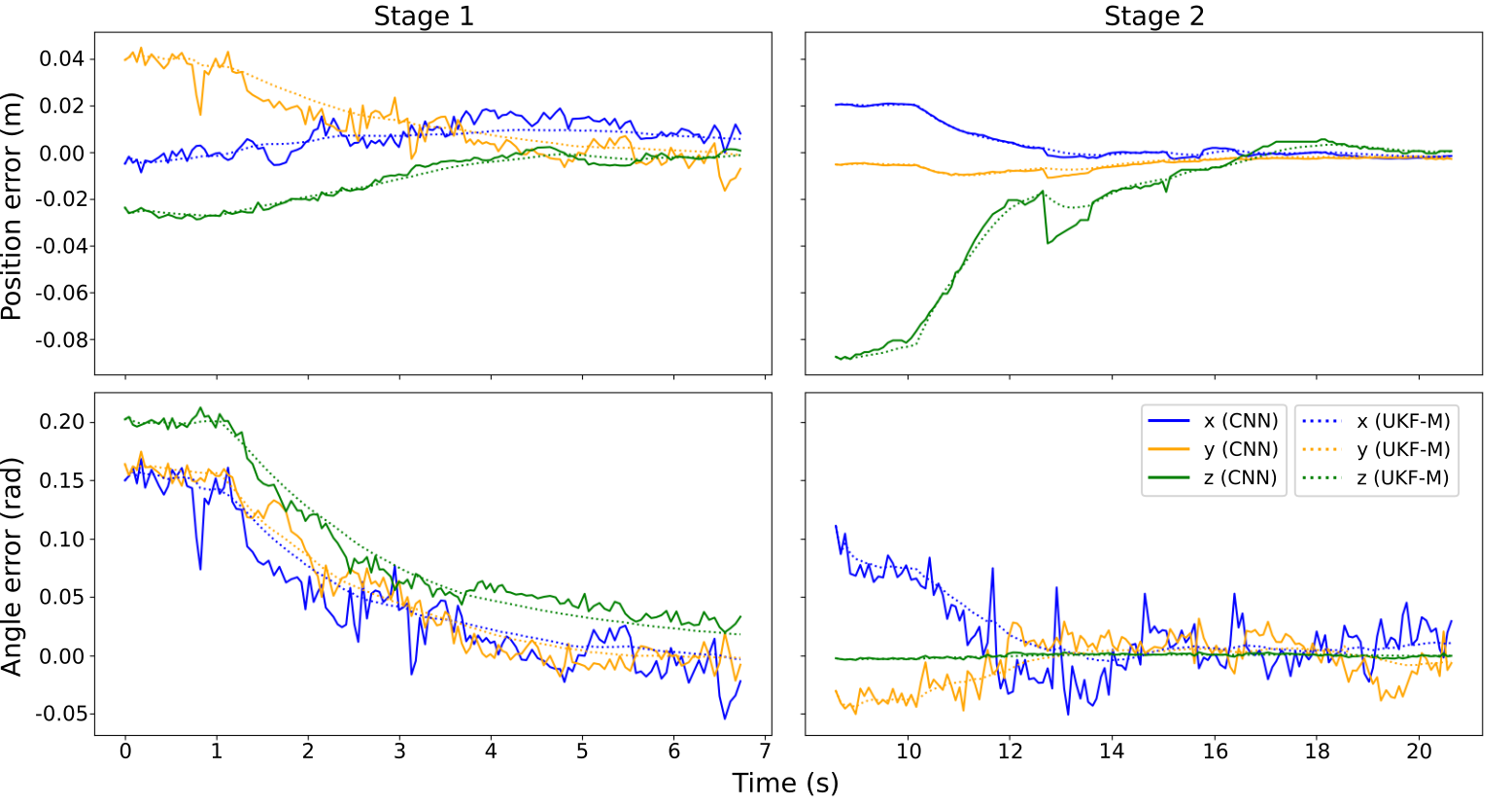}
  \caption[Convergence and comparison of measurements to UKF-M]{Graph showing the convergence of the relative pose along the position (top) and orientation
    (bottom) axes for an example trial. The graph is divided into stage 1 (left) and stage 2
    (right). For stage 2, the Z-axis angular error is low due to only needing to meet two rotational degrees of freedom
    (\ref{eq:min_z_rotation}).}
  \label{fig:vs_convergence}
\end{figure}

As we alluded to in Section \ref{sec:ukfm}, there is substantial noise from the pose measurement,
which is a major reason the UKF-M was included into the system.
Fig. \ref{fig:vs_convergence} shows an example of the relative target pose estimated from the measurements and the UKF-M over
the latter two stages. One can notice that the raw measurements from the CNN
pipeline exhibit substantial amounts of stationary noise in all axes. The estimates from the
UKF-M appear to effectively attenuate the noise and does not lag, which indicates that the process updates (\ref{eq:eq_state_trans}) are
able to contribute to keeping an accurate state. 

\begin{figure}\begin{center}
  \includegraphics[width=0.7\linewidth]{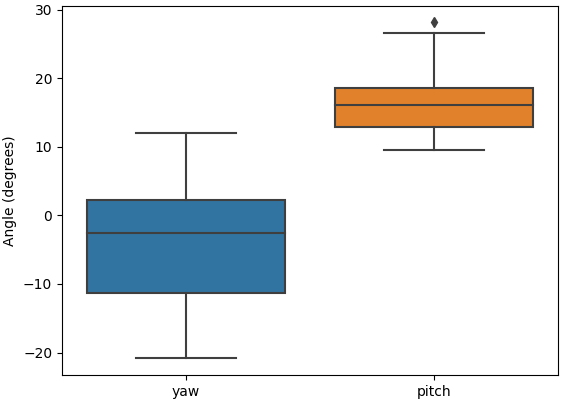}\end{center}
  \caption[Distribution of final observed angles from human trials.]{Distribution of swab angles with respect to the nostril. The horizontal yaw angle generally had wider spread than the vertical pitch angles.}
  \label{fig:obs_swab_ang_distr}
\end{figure}

We examine the level of measurement noise during the trials by subtracting the measurements from the filtered states
for the last two stages during the experiment trials. Fig. \ref{fig:obs_swab_ang_distr} shows the range of standard deviations of the noise
estimate that were recorded during the trials, with two trials excluded due to pauses during the trajectory. For the position noise, stage 2 had an average
standard deviation of 8.2 mm, while stage 3 was lower at 2.3 mm. The angular noise had standard deviations of
0.030 rad ($1.7^\circ$) for stage 2 and 0.017 rad ($1.0^\circ$) for stage 3. Overall, the second stage had more noise, likely
because of the greater distance from the face and higher velocities compared to stage 3.


\subsection{Impact of demographics}
\label{sec:demo}
Given the data driven components nature of 3DDFA\_V2, it is important if the trial outcomes (i.e., whether
the swab reached the nostril, and the final angle) are influenced by participant demographics.
It is clear that the variance in final swab to nose angles exceed any amount that could be explained by the swab to camera pose
calibration. Therefore, the main source of the variance must come from estimates of 3DDFA\_V2. 
Therefore, we examine the potential impact of demographic bias by examining the following three questions:
\begin{enumerate}
\item Are the observed final angles correlated to the individual? If not, then there will be no way
  to establish links between demographic groups. 
\item Does sex significantly impact the final angle or the ability to reach the nostril?
\item Does race  significantly impact the final angle or the ability to reach the nostril?
\end{enumerate}
We apply statistical testing to help  answer these questions. 
Because we are conducting multiple tests,
we will correct for type II errors by applying Bonferroni correction 
and reducing the critical p-value threshold to 0.05/3 = 0.017.

1) For yaw, the standard deviation between different people (intra-person) was 7.1$^\circ$, while the average
standard deviation between the two trials for the same person (inter-person) was only 2.4$^\circ$.
For pitch, the intra-person standard deviation was 4.0$^\circ$ and the inter-person standard deviation was 1.5$^\circ$.
We computed F-statistics by taking the ratio of intra-person variance to inter-person variances, which was
significant for both yaw ($F=5.1, p=9.1 \times 10^{-5}$) and pitch ($F=4.0,
p=4.9 \times 10^{-4}$). 
This presents strong evidence that the angles output by the 3DDFA\_V2 result in repeatable
final orientations with the same face, compared to the final orientations of different faces.

2)  Evaluating the impact of sex, we observe that males have a yaw of -5.3$^\circ$ $\pm$ 6.4$^\circ$ and pitch of 16.5$^\circ$ $\pm$ 4.3$^\circ$, while females had a yaw of
-1.4$^\circ$ $\pm$ 7.8$^\circ$ and pitch of 15.9$^\circ$ $\pm$ 3.1$^\circ$. A t-test reveals that there is no significant angular difference for yaw 
($T=-1.3$, $p =  0.22$) or pitch ($T=0.35$, $p = 0.73$) between the two sexes. Overall, we found that
sex did not play a significant role in the final evaluated angle.
 We found the rate of reaching the nostril was higher for females
(100\%) compared to males (76\%). We applied a Chi-square test to statistically evaluate if the
outcome was biased by sex. Ultimately, this resulted in a test statistic of $\chi^2=3.0$ and
$p=0.08$. Even though the swab tip was able to reach the nostril for all female participants, the difference in performance from males falls below the threshold of statistical significance in our study sample.

3) Evaluating the impact of race from our experiments has a few complications that require some discussion. 
First, facial structure is dependent on many 
other factors than simple racial groupings, which is a topic that is obviously outside
the scope of this article. From our analysis we merely hope to detect if there are any conspicuous biases towards
racial groups that could have been caused by under-representation in the training sets of 3DDFA\_V2. 
The second issue comes from the categorization of races. In the
questionnaire given to participants, we based the categorization from the Government of Canada's census~\cite{Canada2022-408}. 
However, there are many
different races listed,  
and there does not appear to be a standardized way of organizing them into related ethnic groupings. In addition, they did not have a clear
group for certain participants, such as those who were of Iranian descent. Another difficulty concerns how to categorize mixed race participants, which is a scenario
that does not seem to have a universal solution to in the demography literature~\cite{Liebler2008-356}. Ultimately, we subdivided the groups into Group 1=\{Arab, Iranian\}, Group 2 = \{East Asian\}, Group 3 =
\{South Asian, Southeast Asian\}, Group 4 = \{Latin American, White\}, and chose to use ``the inclusive
membership''~\cite{Liebler2008-356} strategy, where mixed race people are included to multiple groups. We applied a one-way
ANOVA model, with an F-test indicating that there is no significant yaw ($F=0.32$, $p=0.82$) or
pitch ($F=1.1$, $p=0.37$) difference between the races. Of course, there are some caveats to this
analysis that we will further elaborate in Section \ref{sec:vs:disc}.
For reaching the nostril, we compare if race is a significant factor by applying a Chi-square
test. With the inclusive membership strategy, a test statistic of $\chi^2 = 2.5$ with $p =
0.47$ indicates that the outcome is not significantly different according to race.

\subsection{Workspace analysis for the contact phase}
\label{sec:workspace}
\begin{figure}\centering
  \includegraphics[width=0.7\linewidth]{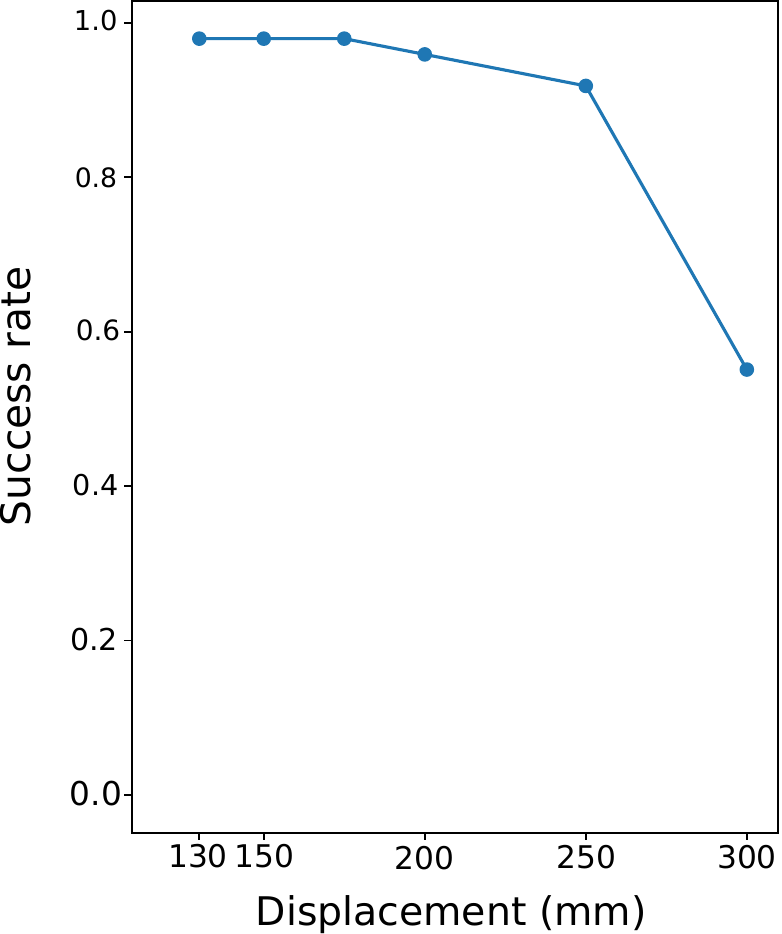}
  \caption[Workspace limits of final joint configurations]{Figure showing how far the end-effector can reach beyond the terminal configuration
    from the trials joint limits and/or collisions are encountered. The final configuration is
    clearly able to achieve the 130 mm displacement that would be needed to reach the nasopharynx
    for the majority of the population. The arm is able to extend several centimeters beyond
    this, indicating that there is sufficient workspace available to proceed with the contact stage
    of the test. }
  \label{fig:success_workspace_extend}
\end{figure}

While the scope of these experiments was to evaluate the visual servo task for NP swab sampling, the arm would need to be capable of executing the contact phase of the test.
To verify this, we will examine the workspace of the arm at the terminal joint configurations from
the trials and test if the swab would still be capable of reaching the nasopharynx without running
into joint limits. The pseudo-inverse Jacobian IK method (\ref{eq:jpinv_IK}) was applied to
integrate joint positions until the tip reaches the target pose, which is 0.2 radians up and at least 130 mm forward. 
We set the threshold for success at a 130 mm because it is longer than a large proportion of the population's nasal cavity~\cite{Liu2009-171}, but
we still evaluated the workspace up to 300 mm. 
Fig. \ref{fig:success_workspace_extend} shows the success rate of the reaching targets of increasing displacement without encountering joint
limits or self-collisions.
All but one trial, where there was excess movement of the participant, was able to reach the target displacement of 
130 mm. In fact, all the trials had sufficient workspace to extend to 175 mm, before the success rate started dropping at 200 mm and
beyond. 
Therefore, we conclude that our joint lookup table is effective at ensuring the arm has sufficient workspace to perform the NP swab test.

\section{Discussion of human trial experiments}
\label{sec:vs:disc}

Overall, we have shown that the proposed system using a single RGB-D camera is effective at performing the pre-contact phase of the NP swab test.
The results from the experiments have highlighted the utility of the UKF-M filtering component within the
proposed system.
We saw that there is significant stationary noise coming from the measurement pipeline, particularly
in terms of the rotation estimates from 3DDFA\_V2. The UKF-M was able to
attenuate this noise effectively and is a necessary component of the pipeline. Without
it, the control response would have to contend with these noisy measurements and be less stable or require the control law to
use lower gain to compensate. Overall the parameters we chose for the UKF-M appear to be appropriate
based on their ability to follow head motion and suppress noise. Of course, there is still some room to modify the chosen parameters. For
example, if we were to assume we
were swabbing a less cooperative cohort, then we would need to increase the magnitude of the measurement
covariance matrix to compensate.

The implemented PBVS control law was effective at placing the swab at the desired pose during trials.
Thanks to the feedback of control velocities to the UKF-M, the pose state stayed current and the task was able to converge
to the desired target without overshooting. One comment from some participants was the duration of the servo component
being too
long. The current control scheme uses a constant gain term that results in an exponential decay in
error over time. This could be improved in the future by implementing an adaptive gain, such as
the method by Kermorgant and Chaumette~\cite{Kermorgant2014-365}, that raises the gain as the error shrinks, allowing for a
quicker, more linear convergence rate.

In terms of the final angle of the swab, we found that the pitch angle was fairly consistent, which would be important for ensuring the swab
does not travel down the wrong tract during the contact phase. The yaw angle were more variable, however, and could be improved.
In addition to biases from 3DDFA\_V2, there could be some additional sources of perceived variance.
The positioning of the bottom camera may have induced some dissimilar perspective effects, especially for taller people. This is a side effect of
having unrestrained participants to evaluate the workspace and motion compensation abilities of the system; in the future this effect could
be examined more critically be repeating the experiment with participants in a head restraint. Another idea could be using a marker-based infrared motion capture
system such as VICON (Yarnton, United Kingdom), but it is not clear if placing markers on a participants face would impact the vision pipeline.

From our statistical analysis, we did not find a significant difference in outcomes among the
demographics of our sample. However, our sample has limitations in terms demographic coverage.
The sample was biased towards the university engineering cohort. This meant that the sex
ratio was skewed towards males (17:8), most of the participants were young adults, and some
racial groups were under-represented or not represented at all. Also, the distribution of our limited sample size across multiple demographic groups reduces statistical power, making it challenging to draw robust conclusions.
Therefore, further evaluation on a wider demographic sample would be needed before application to
clinical settings to ensure the pipeline can generalize appropriately. But, we predict that the
system should generalize well, because 3DDFA\_V2 was trained on a dataset derived from the 300 Faces
in the Wild dataset~\cite{Sagonas2013-375} that consists of images from the internet and wide-reaching social medias, so the model should generalize to a wider demographic than was tested in our
experiments. 
In any case, future studies will need to expand demographic coverage and consider different groupings in statistical analyses.

The arm was able to maintain sufficient workspace during the experiment trials and afterwards if it
were to proceed with the contact phase of the NP swab test. The lookup table provided joint configurations that ensured the arm had
sufficient workspace to execute the task. In terms of further efforts, future implementations could have a detector to indicate if the
participant steps too far away from their initial position after the pre-contact phase begins.

\subsection{Avenues for improvement}
\label{sec:vs_improvements}
\begin{figure}
  \includegraphics[width=0.49\linewidth]{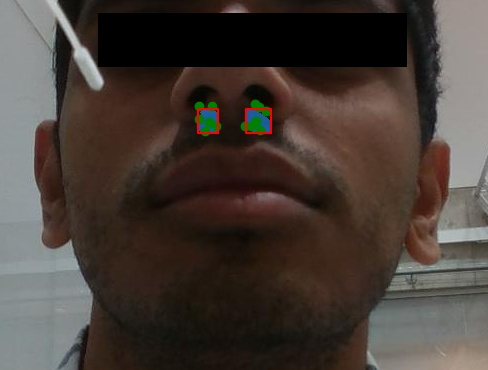}\includegraphics[width=0.49\linewidth]{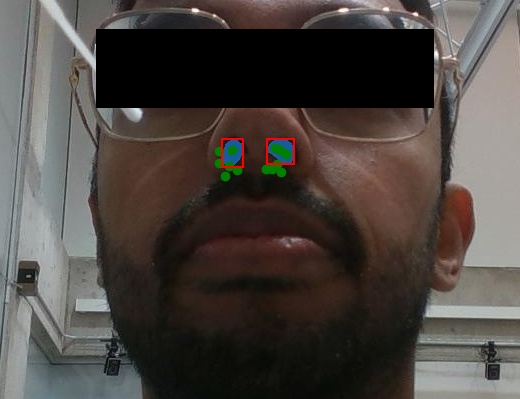}
  \caption[SOT initialization errors]{Images comparing the initial bounding box of the instance where the swab tip missed (left), and
    one that arrived successfully (right). The green dots are the nostril points projected by
    3DDFA\_V2, the light blue mask is the segmentation from SAM~\cite{Kirillov2023-393}, and the right outline is
    the corresponding bounding box. Notice how in the left image none of the nostril points from 3DDFA\_V2 are near
    the nose so it does receive a valid bounding box to begin with.}
  \label{fig:missed_nostril}
\end{figure}

\begin{figure}
  \includegraphics[width=0.49\linewidth]{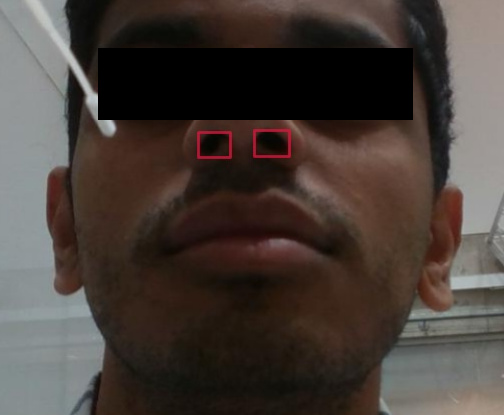}\includegraphics[width=0.49\linewidth]{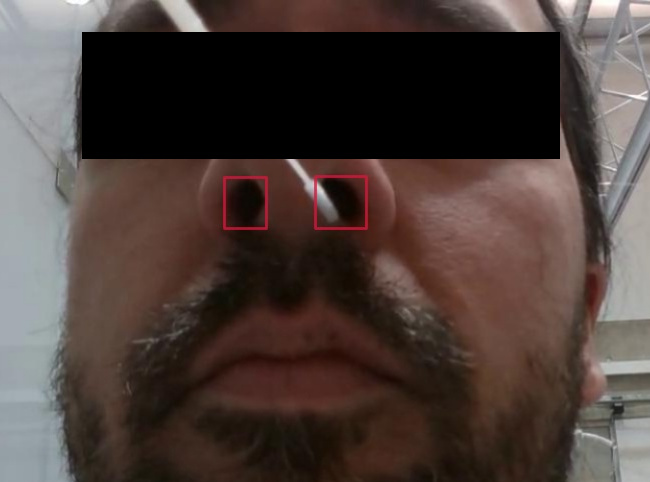}
  \caption[Improved nostril detector]{An improved detector model using a Faster\_RCNN architecture. The detector produces more accurate bounding
    boxes for the nostrils than the previous pipeline (left) and can be tuned to detect the nostril even if it is partially occluded by the swab (right).}
  \label{fig:retrained_model}
  \end{figure}

We saw that in a number of cases the swab tip failed to reach the nostril and settled at another point
near the nose. As mentioned earlier, this was often because the initial bounding box was not aligned
with the nostril because 3DDFA\_V2 predicted the corresponding nostril vertices incorrectly, as
shown in \ref{fig:missed_nostril}. Consequently, the SOT network failed because it did not have a
correct initial bounding box. The pipeline specific to stage 3 is something that we
will
replace in future applications because relying on the unsupervised segmentation and tracking modules appears to be less suitable
when the appearance of the nose and nostril have such distinct visual cues. It would be more sensible to train a dedicated
nostril detector for use in stage 3, similar to Hwang \textit{et al.}~\cite{Hwang2022-310}.  We initially explore this approach by training a CNN to detect the
position of the nostril from the end-effector camera's perspective. The base model is a Faster R-CNN~\cite{Ren2017-399} with a Resnet-50~\cite{He2015-402} backbone that was pretrained on the Coco dataset~\cite{Lin2014-401}. We apply transfer learning by
creating a small dataset using the images from the end-effector camera at various times of the insertion, with 11
participants for training and 14 for testing. 
Notably, the testing set contained all the participants whose nostrils were not reached by the visual servo.
Fig. \ref{fig:retrained_model} shows an example of the detector on two testing images. We trained the model so that it
can identify the nostril even if the swab is partially occluding it. Overall, this approach appears to be reliable and a
good replacement for the unsupervised components.

Like the Eye-to-hand visual servo of Hwang \textit{et al.}~\cite{Hwang2022-310}, the Eye-in-hand visual servo was able to successfully converge to
the nostril. Compared to an Eye-to-hand setup, the Eye-in-hand setup is simple to calibrate
and is not susceptible to needing recalibration from the camera being perturbed because it is rigidly
fixed to the end-effector. The setup is also robust to occlusions because the links
of the robot will never be in a position to obscure the target face. In addition, the swab only needs to
be segmented once, not for each frame like the eye-to-hand setup because the transform between it and the camera remain
constant throughout the trial. However, one drawback of the
eye-in-hand setup is that depth images on most RGB-D cameras, including the RealSense D435i, have a
minimum acquisition range where surfaces closer than this distance cannot be measured. This was not an
issue with the pre-contact phase because there was sufficient space between the swab tip and the
camera. But, if the swab were to enter the nasal cavity, then depth acquisition would eventually be
lost as the camera gets too close to the face. Thus, during the contact phase, adding a camera for a hybrid setup~\cite{Flandin2000-377} could provide some benefit for keeping the swab aligned in conjunction with force inputs. We could also consider complementing the system with additional modalities, such as LiDAR.

\section{Conclusion}
\label{sec:vs:conclusion}
    In this manuscript we proposed a vision guided pipeline to accomplish the pre-contact phase of the NP swab test. The pipeline uses a joint
lookup table to maneuver towards a participant's face with a joint configuration that has sufficient end-effector workspace. Subsequent
stages use deep learning models combined with an unscented Kalman filter on manifolds to estimate the pose of the face, which are
then used to actuate the robot with a pose-based visual servo control loop. We validated our method with a series of human trials, which
were mostly successful and revealed ways to improve nostril tracking to be more reliable and make to make convergence of the control task quicker. In the future, we can consider multi-camera configurations to support vision during the contact phase.

\bibliographystyle{IEEEtran}
\bibliography{ref}

\begin{thebibliography}{10}
\providecommand{\url}[1]{#1}
\csname url@rmstyle\endcsname
\providecommand{\newblock}{\relax}
\providecommand{\bibinfo}[2]{#2}
\providecommand\BIBentrySTDinterwordspacing{\spaceskip=0pt\relax}
\providecommand\BIBentryALTinterwordstretchfactor{4}
\providecommand\BIBentryALTinterwordspacing{\spaceskip=\fontdimen2\font plus
\BIBentryALTinterwordstretchfactor\fontdimen3\font minus
  \fontdimen4\font\relax}
\providecommand\BIBforeignlanguage[2]{{%
\expandafter\ifx\csname l@#1\endcsname\relax
\typeout{** WARNING: IEEEtran.bst: No hyphenation pattern has been}%
\typeout{** loaded for the language `#1'. Using the pattern for}%
\typeout{** the default language instead.}%
\else
\language=\csname l@#1\endcsname
\fi
#2}}

\bibitem{Leber2020-173}
A.~L. Leber, J.~G. Lisby, G.~Hansen, R.~F. Relich, U.~V. Schneider, P.~Granato,
  S.~Young, J.~Pareja, I.~Hannet, and Y.-W. Tang, ``Multicenter evaluation of
  the qiastat-dx respiratory panel for detection of viruses and bacteria in
  nasopharyngeal swab specimens,'' \emph{Journal of Clinical Microbiology},
  vol.~58, no.~5, pp. e00\,155--20, 2020.

\bibitem{Hiebert2021-278}
N.~M. Hiebert, B.~A. Chen, and L.~J. Sowerby, ``Variability in instructions for
  performance of nasopharyngeal swabs across {Canada} in the era of {COVID-19}
  -- what type of swab is actually being performed?'' \emph{Journal of
  Otolaryngology - Head {\&} Neck Surgery}, vol.~50, no.~1, p.~5, Jan 2021.

\bibitem{Fabbris2021-247}
C.~Fabbris, W.~Cestaro, A.~Menegaldo, G.~Spinato, D.~Frezza, A.~Vijendren,
  D.~Borsetto, and P.~Boscolo-Rizzo, ``Is oro/nasopharyngeal swab for
  {SARS-CoV-2} detection a safe procedure? complications observed among a case
  series of 4876 consecutive swabs,'' \emph{Elsevier Inc.}, vol.~42, no.~1, pp.
  102\,758--102\,758, 2021.

\bibitem{Bleier2020-396}
B.~S. Bleier and K.~C. Welch, ``Preprocedural {COVID-19} screening: Do
  rhinologic patients carry a unique risk burden for false-negative results?''
  \emph{International Forum of Allergy \& Rhinology}, vol.~10, no.~10, pp.
  1186--1188, 2020.

\bibitem{12021-435}
{Health Workforce Department - Working Paper 1}, ``The impact of {COVID-19} on
  health and care workers: a closer look at deaths,'' \emph{World Health
  Organization}, no. WHO/HWF/WorkingPaper/2021.1, Sept 2021.

\bibitem{Chen2022-391}
Y.~Chen, Q.~Wang, C.~Chi, C.~Wang, Q.~Gao, H.~Zhang, Z.~Li, Z.~Mu, R.~Xu,
  Z.~Sun, and H.~Qian, ``A collaborative robot for {COVID-19} oropharyngeal
  swabbing,'' \emph{Robotics and Autonomous Systems}, vol. 148, p. 103917,
  2022.

\bibitem{Sun2023-420}
F.~Sun, J.~Ma, T.~Liu, H.~Liu, and B.~Fang, ``Autonomous oropharyngeal-swab
  robot system for {COVID-19} pandemic,'' \emph{IEEE Transactions on Automation
  Science and Engineering}, vol.~20, no.~4, pp. 2469--2478, 2023.

\bibitem{Wang2024-423}
Z.~Wang, S.~Li, X.~He, R.~Chen, L.~Liu, S.~Li, and H.~Liu, ``The {LINGCAI-II}
  system: A sampling robotic system for autonomous oropharyngeal swab
  sampling,'' \emph{IEEE Transactions on Medical Robotics and Bionics}, vol.~6,
  no.~2, pp. 448--459, 2024.

\bibitem{Wang2020-397}
X.~Wang, L.~Tan, X.~Wang, W.~Liu, Y.~Lu, L.~Cheng, and Z.~Sun, ``Comparison of
  nasopharyngeal and oropharyngeal swabs for {SARS-CoV-2} detection in 353
  patients received tests with both specimens simultaneously,''
  \emph{International Journal of Infectious Diseases}, vol.~94, pp. 107--109,
  2020.

\bibitem{Wang2020-196}
S.~Wang, K.~Wang, H.~Liu, and Z.~Hou, ``Design of a low-cost miniature robot to
  assist the {COVID-19} nasopharyngeal swab sampling,'' \emph{CoRR}, vol.
  abs/2005.12679, 2020.

\bibitem{Chen2022-313}
W.~Chen, Z.~Chen, Y.~Lu, H.~Cao, J.~Zhou, M.~C.~F. Tong, and Y.-H. Liu,
  ``Easy-to-deploy combined nasal/throat swab robot with sampling dexterity and
  resistance to external interference,'' \emph{IEEE Robotics and Automation
  Letters}, vol.~7, no.~4, pp. 9699--9706, 2022.

\bibitem{Haddadin2024-390}
S.~Haddadin, D.~Wilhelm, D.~Wahrmann, F.~Tenebruso, H.~Sadeghian, A.~Naceri,
  and S.~Haddadin, ``Autonomous swab robot for naso- and oropharyngeal
  {COVID-19} screening,'' \emph{Scientific Reports}, vol.~14, no.~1, p. 142,
  Jan 2024.

\bibitem{Lee2022-309}
P.~Q. Lee, J.~S. Zelek, and K.~Mombaur, ``Simulating and optimizing
  nasopharyngeal swab insertion paths for use in robotics,'' in \emph{2022 9th
  IEEE RAS/EMBS International Conference for Biomedical Robotics and
  Biomechatronics (BioRob)}, 2022.

\bibitem{Lee2024-436}
------, ``Collaborative robot arm inserting nasopharyngeal swabs with
  admittance control,'' \emph{arXiv preprint arXiv:2408.11688}, 2024.

\bibitem{Hwang2022-310}
G.~Hwang, J.~Lee, and S.~Yang, ``Visual servo control of {COVID-19}
  nasopharyngeal swab sampling robot,'' in \emph{2022 IEEE/RSJ International
  Conference on Intelligent Robots and Systems (IROS)}, 2022.

\bibitem{Guo2020-122}
J.~Guo, X.~Zhu, Y.~Yang, F.~Yang, Z.~Lei, and S.~Z. Li, ``Towards fast,
  accurate and stable {3D} dense face alignment,'' in \emph{Proceedings of the
  European Conference on Computer Vision (ECCV)}, 2020.

\bibitem{Brossard2020-191}
M.~Brossard, A.~Barrau, and S.~Bonnabel, ``A code for unscented kalman
  filtering on manifolds (ukf-m),'' in \emph{2020 IEEE International Conference
  on Robotics and Automation (ICRA)}, 2020.

\bibitem{SaezTrigueros2018-133}
D.~Saez{-}Trigueros, L.~Meng, and M.~Hartnett, ``Face recognition: From
  traditional to deep learning methods,'' \emph{CoRR}, vol. abs/1811.00116,
  2018.

\bibitem{Wu2019-30}
Y.~Wu and Q.~Ji, ``Facial landmark detection: A literature survey,''
  \emph{International Journal of Computer Vision}, vol. 127, no.~2, pp.
  115--142, 02 2019.

\bibitem{Rathod2014-41}
D.~Rathod, A.~Vinay, S.~Shylaja, and S.~Natarajan, ``Facial landmark
  localization-a literature survey,'' \emph{International Journal of Current
  Engineering and Technology}, vol.~4, no.~3, pp. 1901--1907, 2014.

\bibitem{Zadeh2017-56}
A.~{Zadeh}, Y.~C. {Lim}, T.~{Baltrušaitis}, and L.~{Morency}, ``Convolutional
  experts constrained local model for {3D} facial landmark detection,'' in
  \emph{2017 IEEE International Conference on Computer Vision Workshops
  (ICCVW)}, Oct 2017.

\bibitem{Jackson2017-123}
A.~S. {Jackson}, A.~{Bulat}, V.~{Argyriou}, and G.~{Tzimiropoulos}, ``Large
  pose {3D} face reconstruction from a single image via direct volumetric {CNN}
  regression,'' in \emph{2017 IEEE International Conference on Computer Vision
  (ICCV)}, 2017.

\bibitem{Paysan2009-131}
P.~Paysan, R.~Knothe, B.~Amberg, S.~Romdhani, and T.~Vetter, ``A {3D} face
  model for pose and illumination invariant face recognition,'' in \emph{2009
  Sixth IEEE International Conference on Advanced Video and Signal Based
  Surveillance}, 2009.

\bibitem{Howard2019-207}
A.~Howard, M.~Sandler, B.~Chen, W.~Wang, L.~Chen, M.~Tan, G.~Chu, V.~Vasudevan,
  Y.~Zhu, R.~Pang, H.~Adam, and Q.~Le, ``Searching for {MobileNetV3},'' in
  \emph{2019 IEEE/CVF International Conference on Computer Vision (ICCV)}, nov
  2019.

\bibitem{Chaumette2006-326}
F.~Chaumette and S.~Hutchinson, ``Visual servo control. i. basic approaches,''
  \emph{IEEE Robotics \& Automation Magazine}, vol.~13, no.~4, pp. 82--90,
  2006.

\bibitem{Li2023-343}
G.~Li, S.~Zou, and S.~Ding, ``Visual positioning of nasal swab robot based on
  hierarchical decision,'' \emph{Journal of Shanghai Jiaotong University
  (Science)}, vol.~28, no.~3, pp. 323--329, Jun 2023.

\bibitem{Marchand2005-347}
E.~Marchand, F.~Spindler, and F.~Chaumette, ``{ViSP} for visual servoing: a
  generic software platform with a wide class of robot control skills,''
  \emph{IEEE}, vol.~12, no.~4, pp. 40--52, December 2005.

\bibitem{Kuffner2000-291}
J.~Kuffner and S.~LaValle, ``Rrt-connect: An efficient approach to single-query
  path planning,'' in \emph{Proceedings 2000 ICRA. Millennium Conference. IEEE
  International Conference on Robotics and Automation. Symposia Proceedings
  (Cat. No.00CH37065)}, vol.~2, 2000.

\bibitem{Lynch2017-139}
K.~M. Lynch and F.~C. Park, \emph{Modern Robotics: Mechanics, Planning, and
  Control}.\hskip 1em plus 0.5em minus 0.4em\relax Cambridge University Press,
  2017.

\bibitem{Carpentier2019-404}
J.~Carpentier, G.~Saurel, G.~Buondonno, J.~Mirabel, F.~Lamiraux, O.~Stasse, and
  N.~Mansard, ``The {P}inocchio {C}++ library -- {A} fast and flexible
  implementation of rigid body dynamics algorithms and their analytical
  derivatives,'' in \emph{International Symposium on System Integration (SII)},
  2019.

\bibitem{Levinson2020-218}
J.~Levinson, C.~Esteves, K.~Chen, N.~Snavely, A.~Kanazawa, A.~Rostamizadeh, and
  A.~Makadia, ``An analysis of {SVD} for deep rotation estimation,'' in
  \emph{Advances in Neural Information Processing Systems}, vol.~33, 2020.

\bibitem{Li2019-348}
B.~Li, W.~Wu, Q.~Wang, F.~Zhang, J.~Xing, and J.~Yan, ``Siamrpn++: Evolution of
  siamese visual tracking with very deep networks,'' in \emph{Proceedings of
  the IEEE/CVF Conference on Computer Vision and Pattern Recognition (CVPR)},
  June 2019.

\bibitem{Kirillov2023-393}
A.~Kirillov, E.~Mintun, N.~Ravi, H.~Mao, C.~Rolland, L.~Gustafson, T.~Xiao,
  S.~Whitehead, A.~C. Berg, W.-Y. Lo, P.~Doll{\'a}r, and R.~Girshick, ``Segment
  anything,'' \emph{arXiv:2304.02643}, 2023.

\bibitem{Tsai1987-394}
R.~Tsai, ``A versatile camera calibration technique for high-accuracy {3D}
  machine vision metrology using off-the-shelf tv cameras and lenses,''
  \emph{IEEE Journal on Robotics and Automation}, vol.~3, no.~4, pp. 323--344,
  1987.

\bibitem{httpsmathstackexchangecomusers91768jurvandenberg2013-405}
\BIBentryALTinterwordspacing
J.~van den Berg (https://math.stackexchange.com/users/91768/jur-van-den berg).
  (2013) Calculate rotation matrix to align vector $a$ to vector $b$ in 3{D}?
  [Online]. Available: \url{https://math.stackexchange.com/q/476311}
\BIBentrySTDinterwordspacing

\bibitem{Kalman1960-406}
R.~E. Kalman, ``{A New Approach to Linear Filtering and Prediction Problems},''
  \emph{Journal of Basic Engineering}, vol.~82, no.~1, pp. 35--45, 03 1960.

\bibitem{Julier2004-407}
S.~Julier and J.~Uhlmann, ``Unscented filtering and nonlinear estimation,''
  \emph{Proceedings of the IEEE}, vol.~92, no.~3, pp. 401--422, 2004.

\bibitem{279}
\BIBentryALTinterwordspacing
F.~M. Marty, K.~Chen, and K.~A. Verrill, ``How to obtain a nasopharyngeal swab
  specimen,'' \emph{New England Journal of Medicine}, vol. 382, no.~22, p. e76,
  2020. [Online]. Available:
  \url{https://www.nejm.org/doi/full/10.1056/NEJMvcm2010260}
\BIBentrySTDinterwordspacing

\bibitem{Canada2022-408}
\BIBentryALTinterwordspacing
{Statistics Canada}. (2022) Visible minority and population group reference
  guide, census of population, 2021. [Online]. Available:
  \url{https://www12.statcan.gc.ca/census-recensement/2021/ref/98-500/006/98-500-x2021006-eng.cfm}
\BIBentrySTDinterwordspacing

\bibitem{Liebler2008-356}
C.~A. Liebler and A.~Halpern-Manners, ``{A practical approach to using
  Multiple-Race response data: A bridging method for publicuse microdata},''
  \emph{Demography}, vol.~45, no.~1, pp. 143--155, 02 2008.

\bibitem{Liu2009-171}
Y.~Liu, M.~R. Johnson, E.~A. Matida, S.~Kherani, and J.~Marsan, ``Creation of a
  standardized geometry of the human nasal cavity,'' \emph{Journal of Applied
  Physiology}, vol. 106, no.~3, pp. 784--795, 2009.

\bibitem{Kermorgant2014-365}
O.~Kermorgant and F.~Chaumette, ``Dealing with constraints in sensor-based
  robot control,'' \emph{IEEE Transactions on Robotics}, vol.~30, no.~1, pp.
  244--257, 2014.

\bibitem{Sagonas2013-375}
C.~Sagonas, G.~Tzimiropoulos, S.~Zafeiriou, and M.~Pantic, ``300 faces
  in-the-wild challenge: The first facial landmark localization challenge,'' in
  \emph{2013 IEEE International Conference on Computer Vision Workshops}, 2013.

\bibitem{Ren2017-399}
S.~Ren, K.~He, R.~Girshick, and J.~Sun, ``Faster {R-CNN}: Towards real-time
  object detection with region proposal networks,'' \emph{Institute of
  Electrical and Electronics Engineers (IEEE)}, Jun 2017.

\bibitem{He2015-402}
K.~He, X.~Zhang, S.~Ren, and J.~Sun, ``Deep residual learning for image
  recognition,'' \emph{CoRR}, vol. abs/1512.03385, 2015.

\bibitem{Lin2014-401}
T.~Lin, M.~Maire, S.~J. Belongie, L.~D. Bourdev, R.~B. Girshick, J.~Hays,
  P.~Perona, D.~Ramanan, P.~Doll{\'{a}}r, and C.~L. Zitnick, ``Microsoft
  {COCO:} common objects in context,'' \emph{CoRR}, vol. abs/1405.0312, 2014.

\bibitem{Flandin2000-377}
G.~Flandin, F.~Chaumette, and E.~Marchand, ``{Eye-in-hand / eye-to-hand
  cooperation for visual servoing},'' in \emph{{IEEE Int. Conf. on Robotics and
  Automation, ICRA'00}}, vol.~3, 2000.

\end{thebibliography}

\end{document}